\documentclass[lettersize,journal]{IEEEtran}
\usepackage{multicol}
\usepackage{multirow}
\usepackage{hhline}
\usepackage{makecell}
\usepackage{bm}
\usepackage{enumerate}
\usepackage{color}
\usepackage{enumitem}
\usepackage{booktabs}
\usepackage{graphicx}
\usepackage{ulem}
\usepackage{stfloats}
\usepackage{verbatim}
\usepackage{amsmath}
\usepackage{algorithm}
\usepackage{algpseudocode}
\usepackage{footnote}
\usepackage{url}
\usepackage{cite}
\usepackage{amsfonts}
\usepackage{etoolbox}
\usepackage{indentfirst}
\usepackage{epstopdf}
\usepackage{amssymb}
\usepackage{booktabs}
\usepackage{multirow}
\usepackage{colortbl}
\usepackage{bbding}

\usepackage{epsfig}
\usepackage{siunitx}
\usepackage{bbm}
\usepackage{caption}
\usepackage{subcaption}
\usepackage[table,xcdraw]{xcolor}
\algrenewcommand\algorithmicrequire{\textbf{Input:}}
\algrenewcommand\algorithmicensure{\textbf{Output:}}
\begin{document}

\title{Towards Generalizable Deepfake Detection by Primary Region Regularization}

\author{
      Harry~Cheng,~\IEEEmembership{Student Member,~IEEE},
      Yangyang~Guo~\IEEEmembership{Member,~IEEE},
      Tianyi~Wang,~\IEEEmembership{Member,~IEEE},
      Liqiang~Nie,~\IEEEmembership{Senior Member,~IEEE},
      Mohan~Kankanhalli,~\IEEEmembership{Fellow,~IEEE},
}

\maketitle

\begin{abstract}
The existing deepfake detection methods have reached a bottleneck in generalizing to unseen forgeries and manipulation approaches. Based on the observation that the deepfake detectors exhibit a preference for overfitting the specific primary regions in input, this paper enhances the generalization capability from a novel regularization perspective. This can be simply achieved by augmenting the images through primary region removal, thereby preventing the detector from over-relying on data bias. Our method consists of two stages, namely the static localization for primary region maps, as well as the dynamic exploitation of primary region masks. The proposed method can be seamlessly integrated into different backbones without affecting their inference efficiency. We conduct extensive experiments over three widely used deepfake datasets - DFDC, DF-1.0, and Celeb-DF with five backbones. Our method demonstrates an average performance improvement of 6\% across different backbones and performs competitively with several state-of-the-art baselines. 
\end{abstract}


\section{Introduction}
\label{sec:intro}
\IEEEPARstart{T}{he} significant advancement of realistic face synthesis models makes it accessible to alter a person's portrait~\cite{Dual-Generator,df3,faceshifter}. Through the utilization of deepfake techniques, attackers can easily make celebrity pornography products, malicious political speeches, and deceptive government announcements, triggering widespread public concerns. To mitigate the abuse of face forgeries, the development of effective detection methods becomes imperative.

\begin{figure}[t]
    \centering
        \includegraphics[width=0.48\textwidth]{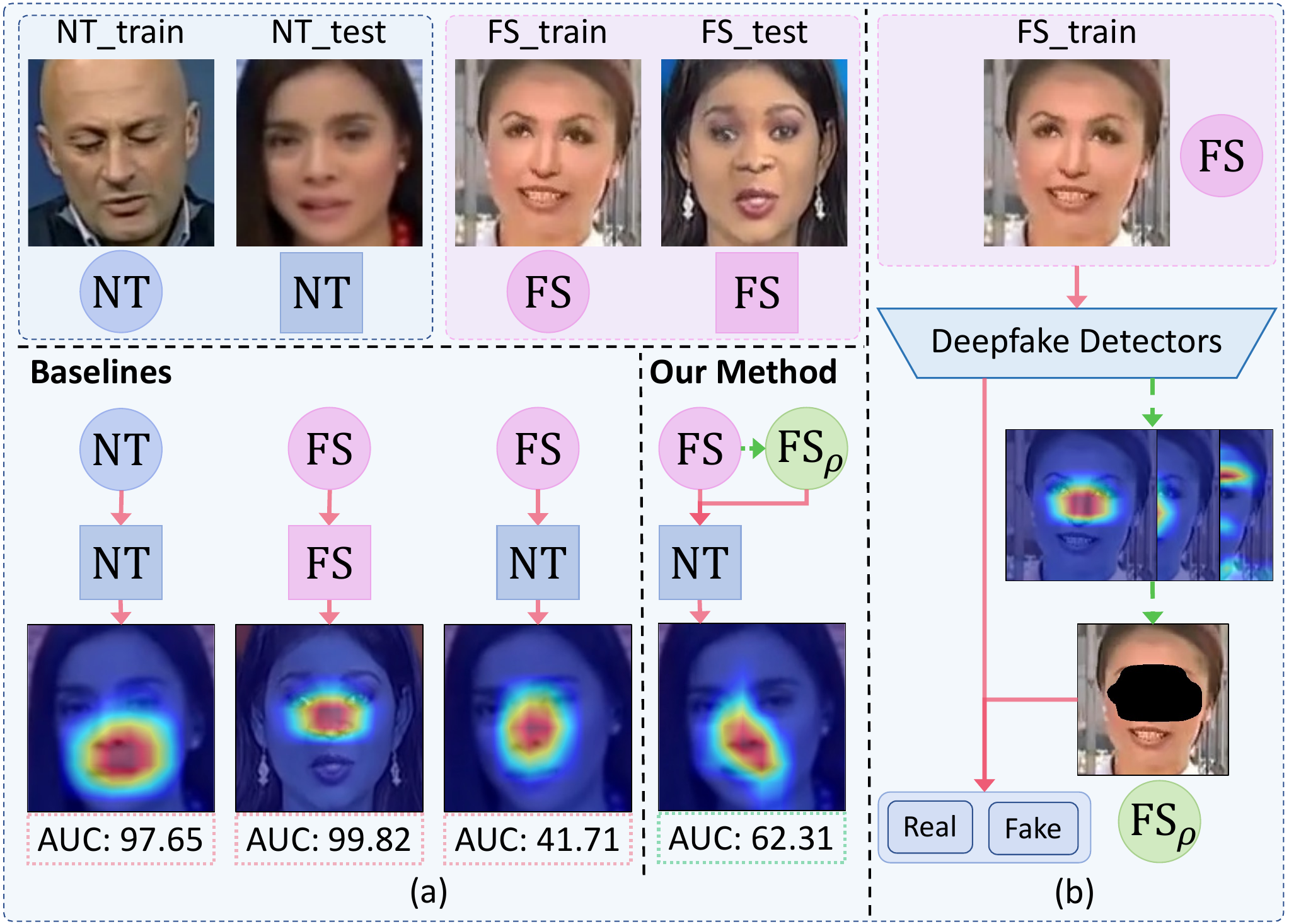}
    \caption{Performance comparison of baselines and ours on NT and FS and our proposed method overview. (a) When trained on NT and FS, the model's attention is respectively placed on lips and noses. Despite the good performance of the baseline over individual datasets, generalization to other datasets is severely hampered (i.e., from FS to NT), where our approach significantly prevails. (b) Our approach augments the training samples to FS$_\rho$ via masking the primary regions. The FS$_\rho$ is used to train the detector with original images jointly.}
    \label{fig:mask_compare} 
\end{figure}

Following a real/fake binary classification paradigm, deepfake detectors~\cite{SRM, F3Net, Learning_Second_Order, Leveraging_Real_Talking} have found considerable success when training and testing on the same datasets, i.e., evaluating under the within-dataset setting. The majority of them identify the deepfakes via discriminating forgery traces, including the manipulated artifacts inside the faces~\cite{lipsdontlie, emotion} and anomalous facial blending patterns~\cite{x-ray, What_Makes_Fake}. 
However, when shifted to unseen datasets or synthetic manipulations, the performance of these models often degrades significantly (refer to Figure~\ref{fig:mask_compare}{a}). 
A dominant reason is the inclination of detectors to overfit specific \textit{primary regions} where the loss function can be optimized most effectively~\cite{RFM}.
This drives the detectors to perform \textit{local} learning while giving up searching for further regions that may be helpful to generalize to unseen data. 
As shown in Figure~\ref{fig:mask_compare}{a}, when trained and tested on the FaceSwap (FS) subset in FF++~\cite{Xception}, the model reaches a 99.82\% AUC score based upon the primary regions that are mostly centered at the \textit{nose}. However, when tested on the NeuralTextures (NT) subset, whose primary regions are \textit{lips}, the model demonstrates significant limitations due to the overfitting of nose regions.
A natural question motivates us - \textit{Is it beneficial to guide the detector to explore beyond the primary regions?}

Existing studies on alleviating the overfitting problem generally follow two directions. 
The first is to leverage differences between real and fake images of the same identity (i.e., the subtraction operation) to guide the detectors to learn fine-grained forgeries~\cite{FakeLocator, facial_manipulated}. These techniques frequently outline the entire face based on the facial contour and skin tone rather than the exact manipulation areas~\cite{On_the_Detection_2020}. 
The other direction is to modify the given faces by erasing random regions~\cite{RFM} or alternative facial attributes~\cite{Face_Reconstruction, chen2022ost}. The additional information brought by this strategy enables models to investigate more hidden features than usual~\cite{What_Makes_Fake}. However, methods in this scope either require random-sized mask generation~\cite{MagDR} or the prior of facial attributes~\cite{Self_ADV, chen2022ost}, which can lead to unstable results. 

This paper addresses the lack-of-generalization problem with a novel view of preventing models from overfitting specific primary regions\footnote{Considering that deepfake methods composite whole faces~\cite{faceshifter}, we believe that there exist cues beyond primary regions that reflect the forgery algorithms.}. 
Specifically, we implement this idea via augmenting the images with the primary regions carefully removed (as shown in Figure~\ref{fig:mask_compare}{b}). 
The augmented data act in the regularization role and help models leverage more clues for detection.
To this end, the key challenge lies in localizing primary regions. 
Inspired by the success of ensemble learning~\cite{Ensemble_1, Ensemble_2}, we propose to employ multiple representative pre-trained deepfake detectors to construct candidate region maps according to the signals from the gradient. 
Based on the consensus of these models, we then design a novel approach to reduce the bias via ensembling the candidate maps to generate a single one. 
Thereafter, we adopt a dynamic exploitation approach to refine the fused map and obtain a more accurate primary region mask to prevent overfitting.
The final masks are overlaid on the original image to obtain the augmented images, which are utilized for training the model with the original ones jointly.


The pipeline automates the acquisition of the primary regions, enabling the training of a robust and generalizable deepfake detection model. Our approach can easily be integrated into different backbones without architectural modifications or adaptations. We conduct extensive experiments on three widely exploited deepfake datasets - DFDC~\cite{dfdc}, DF-1.0~\cite{DF1.0}, and Celeb-DF~\cite{Celeb-DF}. The experimental results demonstrate that the five popular backbones incorporating our method achieve significant improvements in generalization performance, with an average gain of approximately 6\% on AUC under the cross-dataset setting. In addition, our method demonstrates highly competitive results as compared to some SOTA baselines.

In summary, our contributions are three-fold:
\begin{itemize}
\setlength{\itemsep}{2pt}
\setlength{\parsep}{2pt}
\setlength{\parskip}{2pt}
    \item We tackle the deepfake detection from a novel view of regularizing the overfitting of primary regions. With this guidance, our proposed data augmentation method allows models to explore more forgeries by seeing other non-local facial regions. 
    \item We devise a novel region localization strategy to identify the primary regions. Besides enhancing detector generalization abilities, it potentially benefits tasks like forgery trace localization and segmentation.
    \item The experimental results show that the integration of our method will greatly improve the generalizability of backbones, and the performance is comparable to a variety of SOTA competitors.
\end{itemize}

\section{Related Work}
\label{Sec:Rel_Work}
\subsection{Deepfake Generation}
Benefiting from the continuous development of portrait synthesis, deepfake~\cite{DF_PAMI4} has recently emerged as a prevailing research problem. 
The well-studied autoencoders~\cite{autoencoder} serve as the leading architecture in this area~\cite{Face2Face, S_Ob}. Specifically, typical approaches first respectively train two models with the reconstruction task and then swap their decoders to alter the identities of source faces.
These approaches yield realistic faces but are limited to one-to-one face swapping. 
To achieve arbitrary face synthesis, 
Generative Adversarial Networks (GANs)~\cite{GAN} have grown in popularity due to their promising performance. For instance, StyleGAN~\cite{StyleGAN} modifies high-level facial attributes with a progressive growing approach and adaptive instance normalization. IPGAN~\cite{IPGAN} disentangles the identity and attributes of the source and target faces, respectively. These two are thereafter blended for face synthesis. 
Different from these methods, identity-relevant features are recently introduced into deepfake generation.
Kim~\textit{et~al.}~\cite{DBLP:journals/tog/KimCTXTNPRZT18} applied 3DMM~\cite{3DMM2} to produce controllable portraits. 
Xu~\textit{et~al.}~\cite{Region_aware_swapping} augmented local and global identity-relevant features by modeling the cross-scale semantic interaction to achieve identity-consistent face swapping. 

\subsection{Deepfake Detection} 
Deepfake detection~\cite{DF_PAMI2, DF_PAMI3} is generally cast as a binary classification task. Preliminary efforts often endeavor to detect the specific manipulation traces ~\cite{Exploring_Frequency_Adversarial}. Masi~\textit{et~al.}~\cite{Two-Branch} proposed a two-branch network to extract optical and frequency artifacts separately. SSTNet~\cite{SSTNET} detects edited faces through spatial, steganalysis, and temporal features. These models have shown certain improvements on some datasets. Nonetheless, they often encounter inferior performance when applied to different data distributions or manipulation methods.

Several cross-dataset detection approaches are proposed to address this lack-of-generalization issue~\cite{DF_PAMI1}. One manner is to introduce the complementary modalities~\cite{emotion} to vision-only detectors. Zhou~\textit{et al.}~\cite{jointAV} leveraged speech content to detect mismatching mouth-related dynamics. RealForensics~\cite{Leveraging_Real_Talking} exploits the visual and auditory correspondence in real videos to enhance detection performance. Nevertheless, these methods are often limited in certain datasets due to the requirement for additional modalities, e.g., emotion or audio. 
To partially alleviate this limitation, some approaches are proposed to perform data augmentation on the original dataset. For instance, 
FakeLocator~\cite{FakeLocator} calculates the difference between real and fake images to locate the manipulated facial properties.   
Chen~\textit{et al.}~\cite{Self_ADV} specified the blending regions and facial attributes to enrich the deepfake dataset with more manipulation types. 
Wang~\textit{et al.}~\cite{RFM} generated random-sized masks around the pixel with the highest probability of being manipulated. 
These augmentation strategies enable the models to capture more manipulation types and areas during training, thereby enhancing performance.
However, these methods apply either predefined augmentation types or regions, which might trigger bias and lead to sub-optimal outcomes.

\subsection{Deepfake Detection Benchmarks}
The research community has dedicated significant efforts to establishing robust benchmarks for deepfake detection. Early datasets might be relatively small. For instance, DF-TIMIT~\cite{DFTIMIT} contains merely 620 videos synthesized from one manipulation approach. 
Subsequently, larger datasets like FF++~\cite{Xception} incorporate a wide range of manipulation techniques beyond face swapping. 
DFDC~\cite{dfdc} is synthesized from a pool of 960 individuals, resulting in a comprehensive compilation of 100,000 videos. On the other hand, the KoDF dataset~\cite{KODF} comprises over 200,000 videos generated by six distinct algorithms. Recent advancements in deepfake datasets have culminated in increasing sophistication~\cite{Openfor}. For example, FakeAVCeleb~\cite{FakeAVCeleb} modifies audio and video, DGM$^4$~\cite{dgm4} provides detailed forgery grounding annotations, while DF-Platter~\cite{df-platte} involves modifications to multiple faces.
These high-quality datasets encompass diverse data sources and various forgery methods. However, it is crucial to note that this diversity may inadvertently introduce biases between datasets due to unique artifacts from specific forgery generation methods or the use of particular data. Consequently, deepfake detection methods that solely focus on specific artifacts often struggle to generalize effectively across different datasets.

\subsection{Data Augmentation}
Data augmentation is a widely employed technique to combat overfitting in neural networks, which involves modifying the original input to provide the model with diverse and extensive information~\cite{regu_PAMI}.
For instance, Mixup~\cite{mixupfirst} employs linear interpolation between two training samples to enhance generalization based on empirical vicinal risk minimization~\cite{risk}. Cutout~\cite{CutOut} removes contiguous regions from input images, augmenting the training dataset with occluded samples, which helps the model focus on non-dominant parts of the training samples. CutMix~\cite{CutMix} combines two images in the dataset to generate augmented training samples.
Additionally, some studies normalize immediate feature vectors during the training process. 
DropBlock~\cite{DropBlock} drops contiguous regions of feature maps to enhance performance in image classification and object detection tasks. 
PatchUp~\cite{AAAI_regu} improves CNN model robustness against the manifold intrusion problem by applying regularization to selected contiguous blocks of feature maps from a random pair of samples.
Kim~\textit{et al.}~\cite{Kim_regu} proposed feature statistics mixing regularization, which encourages the discriminator's prediction to be invariant to the styles of inputs. Furthermore, Label smoothing\cite{label_smooth} can prevent the model from predicting samples over-confidently via replacing one-hot label vectors with a mixture of labels and uniform distributions.
Our approach adopts the data augmentation method and focuses on the regularization view of primary regions. By directing the models' attention towards non-local information, we aim to improve their generalization capabilities and mitigate the effects of dataset-specific biases.

\section{Methodology}
\label{sec:method}
\subsection{Motivation}
\subsubsection{Preliminaries on deepfake detection}
Most deepfake detection models are trained on datasets that contain both pristine and forged images.
Let $\mathcal{S} = \{(\mathbf{I}_i, y_i)\}_{i=1}^{n}$ be the training set and $\Theta$ be the continuous parameter space, where $\mathbf{I}_i$ is the $i$-th image with respect to the target label $y_i$. For each $\theta \in \Theta$, the empirical risk during training is formulated as:
\begin{equation}
    \hat{R}_\mathcal{S}(\theta):=\frac{1}{n} \sum_{i=1}^n \ell\left(\theta, \mathbf{I}_i, y_i\right),
\end{equation}
where $\ell(\cdot)$ is the loss function such that,
\begin{equation}
    \ell(\theta, \mathbf{I}_i, y_i)=-(y_i \log (\hat{y}_i)+(1-y_i) \log (1-\hat{y}_i)),
\end{equation}
where $\hat{y}_i$ is the score from the predictive function $f_\theta:\mathbf{I} \rightarrow[0,1]$ associated with $\theta$.
To qualitatively demonstrate the ability of a deepfake detector, a widely-used approach is to employ attention maps\footnote{Throughout this paper, the attention maps refer to ones normalized from Grad-CAM rather than those from the attention mechanism.} derived from Class Activation Maps (CAMs)~\cite{Grad-Cam} as the explanation tools~\cite{gradcam_explain, gradcam_explain_2}. 
These attention maps are commonly used to indicate the regions in $\mathbf{I}_i$ that the model focuses on based upon the attention values~\cite{MAT, RFM}. Specifically, the CAM is calculated based on the detector's gradient as follows,
\begin{equation}
\mathbf{A}_i\!=\!\operatorname{ReLU}\left(\sum_{k=1}\left[\frac{1}{UV} \sum_{u=1}^U \sum_{v=1}^V\left(\frac{\partial \hat{y}_i}{\partial \mathbf{H}_{u, v}^k}\right)\right] \mathbf{H}^k\right),
\label{eqn:grad_cam}
\end{equation}
where $\mathbf{H}^k$ is the feature map from the $k$-th channel\footnote{We use the last layer to estimate CAMs following existing work~\cite{VFD}.}, $U$ and $V$ are the width and height of $\mathbf{H}^k$, respectively. 
After normalization, $\mathbf{A}_i$ can be converted into the attention map in the range of [0.0, 1.0], which is then overlaid over $\mathbf{I}_i$ to produce a heat map similar to the ones in Figure~\ref{fig:mask_compare}. 
It is evident that $\mathbf{A}_i$ implies which regions in the feature map are used by detection models to make decisions, and the more important the region is, the larger the attention value is assigned.

\begin{table}[t]
\centering
\caption{The percentage of primary regions to the entire image of several models on different subsets from FF++.}
\scalebox{1.00}{
\begin{tabular}{lccccc}
\toprule \midrule 
\multicolumn{1}{l|}{Model}     & DF & FS  &F2F & Fsh & NT \\ \midrule 
\multicolumn{1}{l|}{Xception~\cite{Xception}}    & 17.79     & 16.62        & 16.72   & 17.50   & 20.79         \\      
\multicolumn{1}{l|}{EfficientNet~\cite{Efficient}}       & 18.40       & 19.25         &  19.80  &  19.86  & 19.58     \\ 
\multicolumn{1}{l|}{VGG~\cite{VGG}}          &  14.86    & 12.64         & 11.88      &  15.89       & 10.64     \\ \midrule \bottomrule   
\end{tabular}}
\label{Tab:primary ratio}
\end{table}

\subsubsection{Overfitting in deepfake detection} 
As shown in the heat maps in Figure~\ref{fig:mask_compare}{a}, the existing detectors tend to overfit a small portion of $\mathbf{I}_i$, i.e., the primary regions. 
These regions can be indicated by $ \mathbf{I}_i* \mathbf{M}_i$, where $*$ denotes the element-wise multiplication, and $\mathbf{M}_i$ is the binary value mask transformed from the attention map $\mathbf{A}_i$,
\begin{equation}
\mathbf{M}_i = \begin{cases}
1, & \text{if } \mathbf{A}_i > 0, \\
0, & \text{otherwise}.
\end{cases}
\label{Eqn:convert}
\end{equation}

It is worth noting that the majority of $\mathbf{M}_i$ values are in fact 0, which means that most image regions contribute nothing to the detection results. As illustrated in Table~\ref{Tab:primary ratio}, over 80\% of the attention values are 0. Using these narrow image areas to detect forgeries leads to overfitting of specific deepfake distribution and harms the generalization ability. This prompts us to make the inference - \textbf{Could the detection generalizability be enhanced by driving the model to focus on regions beyond the primary ones?}




\begin{figure*}[t]
    \centering
        \includegraphics[width=0.93\textwidth]{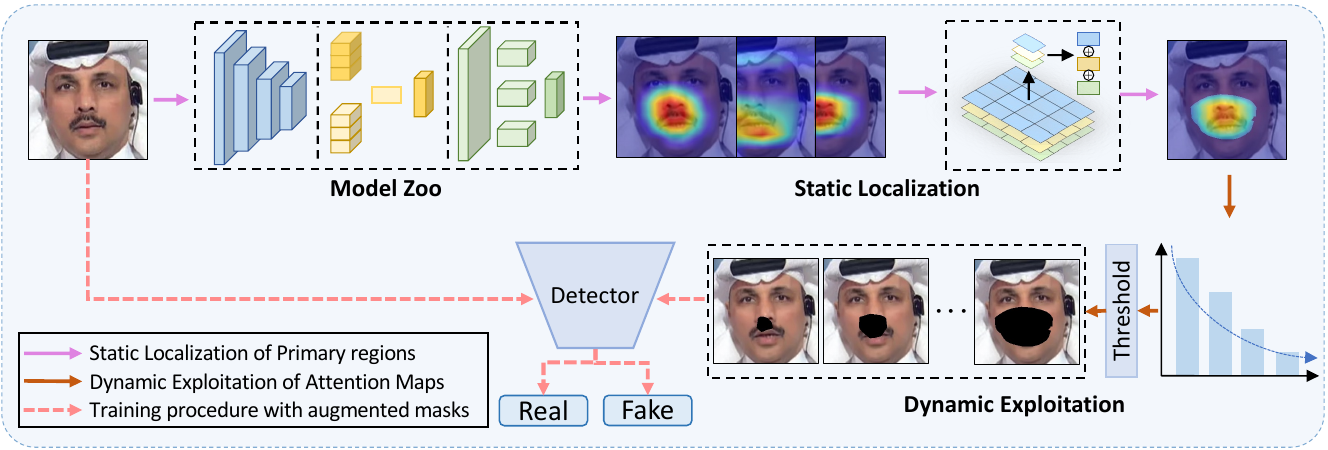}
    \caption{Overview of our PRLE method. We first apply multiple models to localize the candidate primary regions in parallel, where the model bias can be relatively reduced. The exploitation strategy can filter the unnecessary attention areas with a series of masks. Consequently, the augmented data can be effectively employed to existing methods to learn a better detector.}
    \label{fig:overview} 
\end{figure*}

\subsubsection{Regularization perspective} 
Learning a generalizable detector that is versatile in diverse datasets requires seeing beyond these primary regions.
Otherwise, a model often overfits a certain dataset or deepfake algorithm, resulting in sub-optimal performance.
To approach this problem, we propose regularizing the model training to prevent it from being too dependent on primary regions.
This paper implements this idea through the data augmentation technique.
Specifically, we augment the original dataset by carefully removing the primary regions. 
As these new images do not contain the regions that are deemed important by the model, this prevents the model from taking the shortcut and rather forces it to harness other essential cues for decision-making.
In view of this, the augmented data serve as a regularization for the original objective, and the empirical risk is reformulated as:
\begin{equation}
    {R}_\mathcal{S}(\theta):= \hat{R}_\mathcal{S}(\theta) + \frac{1}{n} \sum_{i=1}^n (\ell\left(\theta, \mathbf{I}_i * \overline{\mathbf{M}}_i, y_i\right)),
\label{Eqn: regular}
\end{equation}
where $ \overline{\mathbf{M}}_i = \mathbf{1} - \mathbf{M}_i$ is the complement mask with respect to $\mathbf{M}_i$. Based upon this objective, we propose a \textbf{P}rimary \textbf{R}egion \textbf{L}ocalization and then \textbf{E}xploitation method, dubbed as PRLE in this work. As displayed in Figure~\ref{fig:overview}, our method consists of two stages: static localization and dynamic exploitation. The former obtains joint primary region representations with a novel offline fusion strategy. The latter refines the primary region masks using our proposed augmenting method during the model training.

\subsection{Static Localization of Primary Regions}
\label{sec:AMF}
To achieve the regularization goal, the key is constructing attention maps that can accurately locate primary regions.
Although widely applied as explanation tools, attention maps may exhibit bias due to the dependence on a specific backbone~\cite{Grad_CAM++}. 
We thus apply multiple deepfake detectors with different architectures to obtain attention maps separately and fuse them as a comprehensive map to reduce this bias. 

As shown in Figure~\ref{fig:overview}, we first construct a model zoo $\mathcal{Z} = \{\operatorname{Z}_1, \operatorname{Z}_2, \dots, \operatorname{Z}_T\}$, where $T$ is the zoo size, and $\operatorname{Z}_t \in \mathcal{Z}$ is one pre-trained deepfake detector. Thereafter, we obtain $T$ maps based on Equation~\ref{eqn:grad_cam} and combine them into a map set $\mathcal{A}$ = $\{\mathbf{A}_1, \mathbf{A}_2, \dots, \mathbf{A}_T\}$. To integrate these maps into a single one, an intuitive approach is calculating the average. Specifically, the pixels at the same position $x_i$ in each map from $\mathcal{A}$ are averaged as $s_i$. One threshold $\tau_1$ is then introduced to filter the noise points holding relatively small attention values,
\begin{equation}
\hat{\mathbf{A}}\left(x_i\right)=\left\{\begin{array}{lr}
s_i, & \text{if } s_i>\tau_1, \\
0, & \text{otherwise}.
\end{array}\right.
\label{eqn:tau1_mask}
\end{equation}

\begin{figure}[t]
\centering
\includegraphics[width=0.48\textwidth]{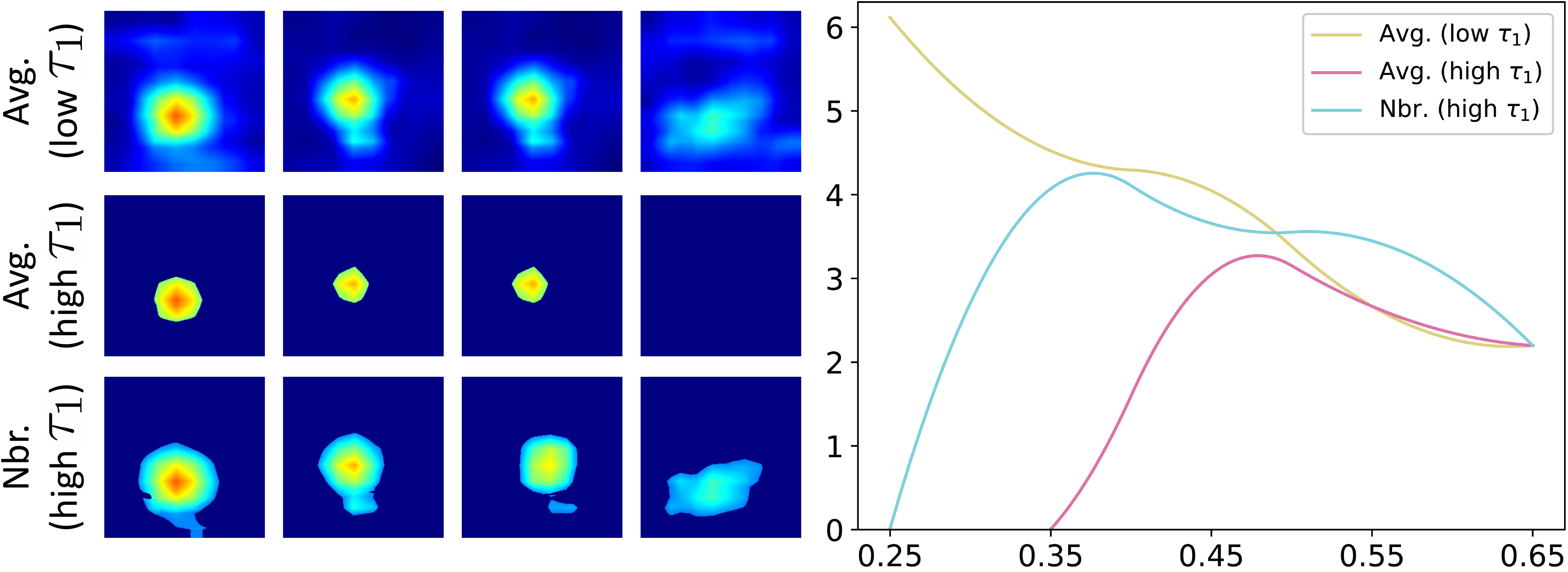}
\caption{Heat maps (left) and data points statistics (right) from the static methods. Left: randomly selected average fused maps with lower $\tau_1$ (1st row) and higher $\tau_1$ (2nd row), and those after applying the neighboring strategy (3rd row). Right: statistical data points with attention values: $x$-axis - attention value; $y$-axis - number of data points (1e7).}
\label{fig:compare_strategy}
\end{figure}

Nevertheless, this operation may cause less satisfactory results. As depicted in Figure~\ref{fig:compare_strategy}, an improper $\tau_1$ can result in excessive noise (first row) or region loss (second row). To address this problem, we design a neighboring fusion method to expand the regions based on the average fused maps with a higher $\tau_1$. In particular, the fused value of a point depends on both itself and its neighboring points,
\begin{equation}
\label{eqn:hor_fusion}
\hspace{-2.5mm}
\operatorname{g}(x_i)\!=\!\mathbbm{1}\!\left\{\!\exists_{\mathbf{A}_j,\mathbf{A}_k \in \mathcal{A}} \! \sum_{x_a \in \mathcal{N}} \! \frac{\!\left|\mathbf{A}_j\!\left(x_a\right)\!-\!\mathbf{A}_k\!(x_i\right)\!|}{|\mathcal{N}|}\!>\!\lambda\!\right\},\!
\end{equation}
where $\mathcal{N}$ is the neighboring set of $x_i$, $\mathbbm{1}$ is the indicator function, and $\lambda$ is a hyperparameter. Equation~\ref{eqn:hor_fusion} indicates that if the existing neighboring point $x_a \in \mathcal{N}$ receives more attention (larger CAM value), we consider $x_i$ to be of interest as well. The value of $\hat{\mathbf{A}}(x_i)$ is taken as the maximum one among its adjacent points, 
\begin{equation}
\hat{\mathbf{A}}(x_i)\!=\!\left\{\begin{array}{lr}
\!\max_{\mathbf{A}_j \in \mathcal{A}, x_n \in \mathcal{N}} \mathbf{A}_j(x_n), \!&\mathrm{g}(x_i)\!=\!1, \\ 
\!0, \!& \mathrm{g}(x_i)\!=\!0. 
\end{array}\right.
\label{eqn:hor_cal}
\end{equation}


As shown in the third row of Figure~\ref{fig:compare_strategy}, after employing the neighboring fusion, the fused maps display more explicit boundaries and maintain the integrity of the primary regions. The statistics of data points exhibit that the neighboring method preserves more data points with high attention than the average one (high $\tau_1$) without adding too much noise. 
This is expected since a high attention value reflects the judgment cues of the models, which are the fundamental indicators of the primary regions. 

The above process can be performed in an offline way. 
Using this, we can identify the common primary regions that models tend to overfit.
Next, we show how we leverage these attention maps for data augmentation.

\begin{figure}[t]
    \centering
        \includegraphics[width=0.48\textwidth]{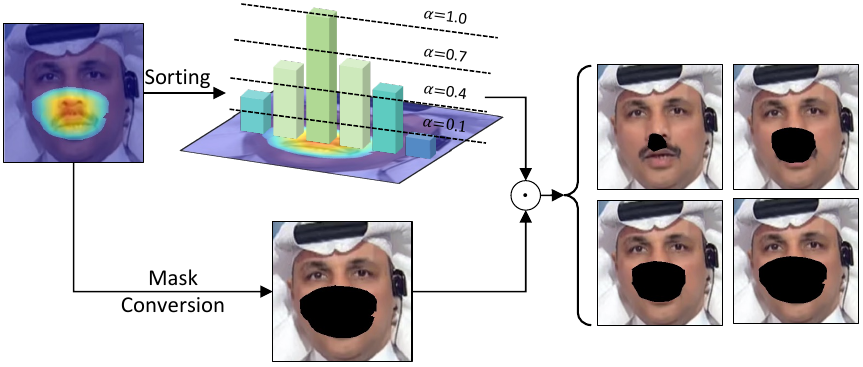}
    \caption{Illustration of our dynamic exploitation stage. The fused heat maps undergo 1) mask conversion, where they are transformed into a binary mask. And 2) a sorting procedure in which the pixels of attention maps are sorted, and only $\alpha$ percentage is reserved. The outputs of these two are combined to make the refined mask images.}
    \label{fig:Finetune_strategy} 
\end{figure}

\subsection{Dynamic Exploitation of Attention Maps}
\label{Sec:refine}
To align the regularization procedure in Equation~\ref{Eqn: regular}, we convert $\hat{\mathbf{A}}$ into the binary mask $\mathbf{M}_b$ using Equation~\ref{Eqn:convert}.
Although it can remove noise and retain the main attention regions, the static stage is prone to localize surplus regions (as seen in Section~\ref{qualititive}). While this can be alleviated by manually adjusting $\lambda$ and $\tau_1$, it requires a burdensome tuning process and results in fixed masks across multiple training epochs that may also cause overfitting. To address this, we propose a dynamic exploitation strategy, where we adjust the size of these masks dynamically during training~\cite{chen2020counterfactual, fine-tune_mask}. 

As illustrated in Figure~\ref{fig:Finetune_strategy}, 
we first arrange the values $\hat{\mathbf{A}}(x_i)$ in descending order,
\begin{equation}
\mathcal{V}=\operatorname{Desc.}(\hat{\mathbf{A}})|_{\hat{\mathbf{A}}(x_i)>0},
\end{equation}
where $\mathcal{V}$ is a set recording the position of pixels according to their attention values. We keep the highest values for they receive the most attention,
\begin{equation}
\mathcal{V}_\alpha = \{x_i \in \mathcal{V}: 0 \leq i \leq \alpha|\mathcal{V}|\},
\end{equation}
where $\mathcal{V}_\alpha$ is the set of selected pixels corresponding to specific $\alpha$. Thereafter, we augment the images $\mathbf{I}_i$ with $\mathcal{V}_\alpha$, 
\begin{equation}
\hat{\mathbf{I}}_i = \mathbf{I}_i * (\mathbf{1} - (\mathcal{V}_\alpha \odot \mathbf{M}_b)) ,
\end{equation}
where $\odot$ can be formulated as, 
\begin{equation}
(\mathcal{V}_\alpha \odot \mathbf{M}_b)(x_i)=\left\{\begin{array}{lr}
\mathbf{M}_b(x_i), & x_i \in \mathcal{V}_\alpha, \\
0, & x_i \notin \mathcal{V}_\alpha. 
\end{array}\right.
\label{Eqn:select_point}
\end{equation}

As illustrated in Figure~\ref{fig:Finetune_strategy}, by setting $\alpha$ randomly, we can dynamically augment images and make masks of varying widths for each training epoch. This method helps diversify the augmented data and prevent the model from overfitting, as shown in our experiments.

\begin{algorithm}[t]
  \caption{Training with PRLE}  
  \label{alg:prop}
  \begin{algorithmic}[1]  
    \Require
      The original images $\mathbf{I}_i$, 
      the attention maps $\hat{\mathbf{A}}_i$, 
      the function $\mathrm{Dynamic}$ as described in Section~\ref{Sec:refine},
      and the predictive function $f_\theta:\mathbf{I} \rightarrow[0,1]$; 
   \Ensure
      The predictive score $\hat{y}_i$; 
    \If{$\mathrm{Rand} (0,1) \leq p$}
        \State $\mathbf{I}^{*} \gets \mathrm{Dynamic} (\mathbf{I}_i, 
        \hat{\mathbf{A}}_i, \alpha)$; \\
        \Comment{$\alpha$ is randomly chosen from (0, 1];}
    \Else
        \State $\mathbf{I}^{*} \gets \mathbf{I}_i$;
    \EndIf
    \State $\hat{y}_i \gets f_\theta(\mathbf{I}^{*})$;
    \State \Return  $\hat{y}_i$;
\end{algorithmic}
\end{algorithm}
\vspace{-1em}

\subsection{Training Protocols}
\label{Sec:training}
The static localization of primary regions and their dynamic exploitation are sequentially utilized to curate the augmented image. Specifically, we treat the former stage in the way of data preprocessing using pre-trained models. The latter stage dynamically augmented the training data throughout the training procedure using random $\alpha$. 
\textbf{Notably, we do not double the training set as other augmentation approaches do.} Instead, we choose to either use the original image or its counterpart with primary regions being masked for each training epoch, as described in Algorithm~\ref{alg:prop}. 
In this way, the training burden of the model is increased imperceptibly (Please refer to Section~\ref{Sec:effi} for the analysis of efficiency), and the random setting also prevents the models from overfitting.
The training objective can be formalized as, 
\begin{equation}
\label{regular}
    {R}_\mathcal{S}(\theta)\!:=\! 
    \frac{1}{n} \sum_{i=1}^n\!(q\ell\left(\theta, \mathbf{I}_i, y_i\right) + 
    (1\!-q)(\gamma \ell\left(\theta, \hat{\mathbf{I}}_i, y_i\right)),
\end{equation}
where $q \in \{0,1\}$ indicates whether the input is augmented, and $\gamma$ is a hyperparameter that controls the regularization terms. 
Similar to other regularization methods such as mixup learning~\cite{Mixup}, in which training is performed on synthetic samples from the original training set, our approach approximates regularized loss minimization, making vanilla detectors more generalizable and robust~\cite{Mixupthe, Mixup19}. 
With the constraint of this regularization, detectors have to take into account both primary regions and beyond that can be useful for generalization.

\begin{table*}[]
\centering
\caption{Performance comparison between SOTA baselines (unmarked), backbones (pink background), and backbones with our PRLE method added (marked blue). All the models are trained on the FF++ dataset. The best performance is marked as bold. $^{\ddag}$: the backbone is used to generate the primary region maps.}
\begin{tabular}{l|ccccccccc}
\toprule \midrule
\multirow{2}{*}{{Method}}         &\multirow{2}{*}{{AUG}}      & \multicolumn{2}{|c}{DFDC} & \multicolumn{2}{c}{DF-1.0} & \multicolumn{2}{c}{Celeb-DF} & \multicolumn{2}{|c}{AVG} \\ \cmidrule(lr){3-4} \cmidrule(lr){5-6} \cmidrule(lr){7-8} \cmidrule(lr){9-10} 
                            &     & \multicolumn{1}{|c}{ACC}         & AUC        & ACC           & AUC          & ACC          & AUC     & \multicolumn{1}{|c}{ACC}          & AUC    \\ \midrule 
\multicolumn{1}{l|}{MesoNet~\cite{mesonet}}    &\multicolumn{1}{c|}{$\times$}       & 50.02       & 50.16    & 50.05         & 50.21  & 36.73     & 50.01   & \multicolumn{1}{|c}{45.60}        & 50.13  \\
\multicolumn{1}{l|}{Capsule~\cite{capsule}}    &\multicolumn{1}{c|}{$\times$}     & 51.30       & 56.16     & 59.29         & 61.46  & 61.96     & 59.93   & \multicolumn{1}{|c}{57.52}        & 59.18   \\
\multicolumn{1}{l|}{CViT~\cite{CViT}}      &\multicolumn{1}{c|}{$\times$}        & 60.76       & 67.43       & 54.97        & 58.52     & 53.26     & 63.60     & \multicolumn{1}{|c}{56.33}        & 63.18   \\ 
\multicolumn{1}{l|}{FFD~\cite{On_the_Detection_2020}} &\multicolumn{1}{c|}{$\checkmark$}    & 59.44       & 59.47       & 53.69  & 53.81    & 46.19     & 55.86  & \multicolumn{1}{|c}{53.11}        & 56.38  \\
\multicolumn{1}{l|}{MAT~\cite{MAT}}      &\multicolumn{1}{c|}{$\checkmark$}     & 63.16      & 69.06      & 56.90         & 61.72     & 44.78     & 57.20     & \multicolumn{1}{|c}{54.95}        & 62.66   \\ 
\multicolumn{1}{l|}{SRM~\cite{SRM}}       &\multicolumn{1}{c|}{$\checkmark$}         & 59.95      & 64.80    & 55.83         & 62.54        & 52.95     & 60.90     & \multicolumn{1}{|c}{56.24}        & 62.75   \\ 
\multicolumn{1}{l|}{RFM~\cite{RFM}}      &\multicolumn{1}{c|}{$\checkmark$}     & 60.55      & 66.03       & 59.33         & 60.27     & 62.04     & 65.79   & \multicolumn{1}{|c}{60.64}        & 64.03   \\ 
\multicolumn{1}{l|}{RECCE~\cite{Face_Reconstruction}} &\multicolumn{1}{c|}{$\checkmark$}   & 59.30      & 62.82      & 56.02    & 60.41     & 68.49    & 69.80   & \multicolumn{1}{|c}{61.27}        & 64.34   \\  \midrule 
\rowcolor[HTML]{F4E5F1}\multicolumn{1}{l|}{$^{\ddag}$Xception~\cite{Xception}}  &\multicolumn{1}{c|}{$\times$}  & 59.93  & 64.17   & 48.04   & 55.01  & 56.12  & 56.75   & \multicolumn{1}{|c}{54.70}        & 58.64  \\  
\rowcolor[HTML]{F4E5F1}\multicolumn{1}{l|}{$^{\ddag}$EfficientNet~\cite{Efficient}} &\multicolumn{1}{c|}{$\times$}  & 60.63    & 65.43   & 54.97   & 58.59  & 62.79  & 64.59   & \multicolumn{1}{|c}{59.46}        & 62.87  \\
\rowcolor[HTML]{F4E5F1}\multicolumn{1}{l|}{$^{\ddag}$VGG~\cite{VGG}} &\multicolumn{1}{c|}{$\times$}   & 58.06       & 61.60    & 63.96   & 65.59    & 66.62     & 63.97    & \multicolumn{1}{|c}{62.75}        & 63.72   \\ 
\rowcolor[HTML]{F4E5F1}\multicolumn{1}{l|}{F$^3$-Net~\cite{F3Net}}  &\multicolumn{1}{c|}{$\checkmark$}    & 63.76       & 67.59    & 56.93     & 59.15   & 58.58   & 64.76   & \multicolumn{1}{|c}{59.76}        & 63.83     \\
\rowcolor[HTML]{F4E5F1}\multicolumn{1}{l|}{ResNet-50~\cite{ResNet}} &\multicolumn{1}{c|}{$\times$}   & 59.84    & 64.34   & 58.15     & 62.78    & 56.68    & 60.32   & \multicolumn{1}{|c}{58.22}      & 62.48   \\ \midrule 
\rowcolor[HTML]{D9EEF2}\multicolumn{1}{l|}{$^{\ddag}$Xception + PRLE} &\multicolumn{1}{c|}{$\checkmark$}      & 64.33   & 69.38  & 57.22    & 66.62      & 65.19     & 65.94    & \multicolumn{1}{|c}{62.28}        & 67.31 \\
\rowcolor[HTML]{D9EEF2}\multicolumn{1}{l|}{$^{\ddag}$EfficientNet + PRLE} &\multicolumn{1}{c|}{$\checkmark$}   & \textbf{64.52}   & \textbf{69.64}  & 62.14    & \textbf{74.72}  & \textbf{69.71}   & \textbf{70.67}     & \multicolumn{1}{|c}{\textbf{65.46}}      & \textbf{71.68}  \\
\rowcolor[HTML]{D9EEF2}\multicolumn{1}{l|}{$^{\ddag}$VGG + PRLE} &\multicolumn{1}{c|}{$\checkmark$}       & 56.87       & 62.54      & \textbf{64.92}         & 68.73        & 67.87        & 68.77      &\multicolumn{1}{|c}{ 63.22}        & 66.68 \\ 
\rowcolor[HTML]{D9EEF2}\multicolumn{1}{l|}{F$^3$-Net + PRLE} &\multicolumn{1}{c|}{$\checkmark$}       & 63.50     & 69.47    &  63.32      & 74.30     &  68.46     &  68.17    & \multicolumn{1}{|c}{65.09}        & 70.65  \\
\rowcolor[HTML]{D9EEF2}\multicolumn{1}{l|}{ResNet-50 + PRLE}  &\multicolumn{1}{c|}{$\checkmark$}      & 61.50       & 66.02      & 57.43         & 70.52        & 58.94        & 61.26      & \multicolumn{1}{|c}{59.29}        & 65.93 \\ \midrule 
\bottomrule
\end{tabular}

\label{SOTA}
\end{table*}

\section{Experiments}
\label{sec:expo}
\subsection{Datasets and Baselines}
\noindent\textbf{Training datasets.} Following the common setting of cross-dataset deepfake detection~\cite{RFM, Learning_Second_Order, Face_Reconstruction}, we trained our model on the FF++ dataset~\cite{Xception}. It contains 1,000 original pristine videos collected from YouTube and five types of manipulation techniques, i.e., Deepfakes (DF), FaceSwap (FS), Face2Face (F2F), FaceShifter (Fsh), and NeuralTextures (NT), resulting in 6,000 videos in total. 

\noindent\textbf{Testing datasets.} Three widely used deepfake datasets are applied to evaluate the generalizability of our model. 

\noindent\textbf{1) Deepfake Detection Challenge (DFDC)}~\cite{dfdc} is one of the largest public deepfake datasets by far, with 23,654 real videos and 104,500 fake videos synthesized using eight different facial manipulation methods.

\noindent\textbf{2) DF-1.0}~\cite{DF1.0} includes videos with manually adding deliberate distortions and perturbations to the clean deepfake videos. We followed the official split and performed deepfake detection on the 1,000 testing set videos with mixed distortions.

\noindent\textbf{3) Celeb-DF}~\cite{Celeb-DF} is one of the most challenging deepfake detection datasets. We considered the Celeb-DF official testing dataset with 518 videos in the experiments. 

\subsection{Implementation}
We utilized the Dlib library\footnote{http://dlib.net/.} to extract and align faces, which are then resized to 256 $\times$ 256 for both training and testing sets. These faces are employed for the image-level experiments conducted on two RTX 3090 GPUs with a batch size of 64. To obtain the primary regions, we employed three reliable deepfake detection models with diverse architectures, namely Xception~\cite{Xception}, EfficientNet~\cite{Efficient}, and VGG~\cite{VGG}, where the latter two are pre-trained on ImageNet~\cite{imagenet}. The values of threshold $\tau_1$ and $\lambda$ used in Section~\ref{sec:AMF} are set as 0.3 and 0.15, respectively. We selected the value of $\alpha$ in Section~\ref{Sec:refine} from the range of 0.0 to 1.0, and set $p$ in Algorithm~\ref{alg:prop} and $\gamma$ in Section~\ref{Sec:training} to 0.5 and 1.0, respectively.

\subsection{Performance Comparison}
\subsubsection{Comparison on testing datasets}
We compared the generalization performance of several SOTA baselines, backbones, and our proposed method. 
All the models are trained on FF++ and evaluated on the three testing datasets. This cross-dataset setup is challenging since neither the testing pristine/forged videos nor the manipulated techniques are visible in the training dataset. 
We utilized two metrics to quantify the models' performance: ACC (accuracy) and AUC (area under the receiver operating characteristic curve). Also, the mean AUCs and ACCs over testing datasets are calculated to evaluate the overall performance.
The experimental results are summarized in Table~\ref{SOTA}. 

\noindent\textbf{Comparison with backbones.} 
We first validated the generalizability brought by PRLE via applying it to different backbones. 
Pertaining to the five backbones, i.e., Xception, EfficientNet, VGG, F$^3$-Net, and ResNet-50, the model zoo uses the former three to generate the primary region maps\footnote{As PRLE is used in training set only, this does not lead to label leaking or cheating in the testing stage.}. 
The latter two are applied to validate the applicability of primary region masks, as they have not been utilized previously.

As can be observed, adopting our PRLE approach (marked as \textit{blue}) significantly improves the generalizability of backbones (marked as \textit{pink}). 
For example, the ACC of Xception on the Celeb-DF dataset increases by 9\%, and the AUC of EfficientNet on DF-1.0 gains by almost 15\%, demonstrating that our PRLE drives the detectors to explore additional information beyond the primary regions for generalization to unseen distributions and manipulation.
In addition to the three models involved in the primary maps generation, F$^3$-Net, and ResNet-50 also exhibit significant performance improvement over respective backbones. Both models achieve an average AUC improvement of approximately 3\%. In a nutshell, by simply adding our regularization term, classic detectors can be made more transferable.

\begin{table}[]
\centering
\caption{AUC (\%) comparison between backbones and the models incorporating our PRLE (+P.). Each row represents the model performance trained on a specific subset and tested on all of the five FF++ subsets.}
\scalebox{0.87}{
\begin{tabular}{lcccccccc}
\toprule \midrule
\multicolumn{2}{l|}{Method}   & \multicolumn{1}{c|}{}  \hspace{0.2pt}    & \multicolumn{1}{c}{}    &\multicolumn{1}{c}{}  &\multicolumn{1}{c}{}  &\multicolumn{1}{c}{}  &\multicolumn{1}{c}{}\\ \cmidrule{1-2}
\multicolumn{1}{l|}{Backbone}  & \multicolumn{1}{c|}{+P.}  & \multicolumn{1}{c|}{\multirow{-2}{*}{Train}} \hspace{0.2pt}   & \multicolumn{1}{c}{\multirow{-2}{*}{DF}}    & \multicolumn{1}{c}{\multirow{-2}{*}{FS}}    & \multicolumn{1}{c}{\multirow{-2}{*}{F2F}}   & \multicolumn{1}{c}{\multirow{-2}{*}{Fsh}}   & \multicolumn{1}{c}{\multirow{-2}{*}{NT}}  & \multicolumn{1}{c}{\multirow{-2}{*}{AVG}} \\ \midrule
\multicolumn{1}{l|}{Xception}  & \multicolumn{1}{c|}{$\times$}  & \multicolumn{1}{c|}{\multirow{10}{*}{DF}} \hspace{0.2pt}  & \multicolumn{1}{c}{{\cellcolor[HTML]{cdcbcb}99.91}} & 28.91 & 74.59 & 66.34 & 83.74     & 70.70 \\
\multicolumn{1}{l|}{Xception}  & \multicolumn{1}{c|}{$\checkmark$}   & \multicolumn{1}{c|}{} \hspace{0.2pt}       & \cellcolor[HTML]{cdcbcb}99.89 & \textbf{30.25} & \textbf{80.17} & \textbf{71.90} & \textbf{84.02} & \textbf{73.25} \\
\multicolumn{1}{l|}{EfficientNet}  & \multicolumn{1}{c|}{$\times$}   & \multicolumn{1}{c|}{} \hspace{0.2pt}       & \cellcolor[HTML]{cdcbcb}99.95 & 36.66 & 69.69 & \textbf{69.10} & 81.25                            & 71.33  \\
\multicolumn{1}{l|}{EfficientNet}  & \multicolumn{1}{c|}{$\checkmark$}   & \multicolumn{1}{c|}{} \hspace{0.2pt}  & \cellcolor[HTML]{cdcbcb}98.59 & \textbf{38.28} & \textbf{75.82} & 68.89 & \textbf{85.32}           & \textbf{73.38} \\
\multicolumn{1}{l|}{VGG}    & \multicolumn{1}{c|}{$\times$}     & \multicolumn{1}{c|}{} \hspace{0.2pt} & \cellcolor[HTML]{cdcbcb}99.67 & 26.60          & 63.40          & \textbf{66.34}            & \textbf{73.96} & 65.99 \\
\multicolumn{1}{l|}{VGG}  & \multicolumn{1}{c|}{$\checkmark$}   & \multicolumn{1}{c|}{} \hspace{0.2pt}                  & \cellcolor[HTML]{cdcbcb}99.59 & \textbf{34.63}          & \textbf{65.44}  & 65.29   & 70.74 & \textbf{67.13} \\
\multicolumn{1}{l|}{F$^3$-Net}  & \multicolumn{1}{c|}{$\times$}   & \multicolumn{1}{c|}{} \hspace{0.2pt}                & \cellcolor[HTML]{cdcbcb}99.62 & 21.67          & 60.37          &  70.98           & 76.06  & 65.75 \\
\multicolumn{1}{l|}{F$^3$-Net}  & \multicolumn{1}{c|}{$\checkmark$}   & \multicolumn{1}{c|}{} \hspace{0.2pt}  & \cellcolor[HTML]{cdcbcb}99.64 & \textbf{24.42}   & \textbf{67.51} & \textbf{75.54}  & \textbf{79.19}  & \textbf{69.26}  \\
\multicolumn{1}{l|}{ResNet-50}  & \multicolumn{1}{c|}{$\times$}   & \multicolumn{1}{c|}{} \hspace{0.2pt}                & \cellcolor[HTML]{cdcbcb}99.88 & 24.99          & 65.74          & 71.20    & 74.91 & 67.34  \\
\multicolumn{1}{l|}{ResNet-50}  & \multicolumn{1}{c|}{$\checkmark$}   & \multicolumn{1}{c|}{} \hspace{0.2pt}            & \cellcolor[HTML]{cdcbcb}98.58 & \textbf{26.65}  & \textbf{68.69} & \textbf{72.48}  & \textbf{75.22}  & \textbf{68.32}  \\ \midrule

\multicolumn{1}{l|}{Xception}  & \multicolumn{1}{c|}{$\times$}  & \multicolumn{1}{c|}{\multirow{10}{*}{FS}} \hspace{0.2pt}   & 61.29 & \cellcolor[HTML]{cdcbcb}99.87 & 66.04 & 52.55 & 55.82                          & 67.11    \\
\multicolumn{1}{l|}{Xception}  & \multicolumn{1}{c|}{$\checkmark$}   & \multicolumn{1}{c|}{} \hspace{0.2pt}  & \textbf{62.98} & \cellcolor[HTML]{cdcbcb}99.89 & \textbf{69.30} & \textbf{52.84} & \textbf{62.31}      & \textbf{69.46}      \\
\multicolumn{1}{l|}{EfficientNet}  & \multicolumn{1}{c|}{$\times$}   & \multicolumn{1}{c|}{} \hspace{0.2pt}  & 66.02 & \cellcolor[HTML]{cdcbcb}99.91 & 67.07 & 53.38 & \textbf{54.63}                                 & 68.20     \\
\multicolumn{1}{l|}{EfficientNet}  & \multicolumn{1}{c|}{$\checkmark$}   & \multicolumn{1}{c|}{} \hspace{0.2pt}  & \textbf{66.46} & \cellcolor[HTML]{cdcbcb}99.71 & \textbf{72.90} & \textbf{56.58} & 53.72           & \textbf{69.87}    \\ 
\multicolumn{1}{l|}{VGG}  & \multicolumn{1}{c|}{$\times$}     & \multicolumn{1}{c|}{} \hspace{0.2pt} & 54.35          & \cellcolor[HTML]{cdcbcb}99.64 & 59.71          & 56.75            & 52.24                     & 64.53    \\
\multicolumn{1}{l|}{VGG}  & \multicolumn{1}{c|}{$\checkmark$}   & \multicolumn{1}{c|}{} \hspace{0.2pt}   & \textbf{59.37} & \cellcolor[HTML]{cdcbcb}99.82 & \textbf{60.23} & \textbf{57.74} & \textbf{52.68}          & \textbf{65.96}   \\
\multicolumn{1}{l|}{F$^3$-Net}  & \multicolumn{1}{c|}{$\times$}   & \multicolumn{1}{c|}{} \hspace{0.2pt}                & 37.49          & \cellcolor[HTML]{cdcbcb}99.43 & 68.83          & 46.00            & 39.15  & 58.18     \\
\multicolumn{1}{l|}{F$^3$-Net}  & \multicolumn{1}{c|}{$\checkmark$}   & \multicolumn{1}{c|}{} \hspace{0.2pt}  & \textbf{43.09} & \cellcolor[HTML]{cdcbcb}99.81 & \textbf{72.53} & \textbf{47.19}   & \textbf{43.74}   & \textbf{61.27}   \\        
\multicolumn{1}{l|}{ResNet-50}  & \multicolumn{1}{c|}{$\times$}   & \multicolumn{1}{c|}{} \hspace{0.2pt}                & 45.63 & \cellcolor[HTML]{cdcbcb}99.86 & 60.47          & 49.39            & 52.71           & 61.61    \\
\multicolumn{1}{l|}{ResNet-50}  & \multicolumn{1}{c|}{$\checkmark$}   & \multicolumn{1}{c|}{} \hspace{0.2pt}   & \textbf{57.91}   & \cellcolor[HTML]{cdcbcb}99.84 & \textbf{72.64}  & \textbf{56.82}  &\textbf{62.31} & \textbf{69.90}     \\    \midrule

\multicolumn{1}{l|}{Xception}  & \multicolumn{1}{c|}{$\times$}  & \multicolumn{1}{c|}{\multirow{10}{*}{F2F}} \hspace{0.2pt}  & 82.92 & 50.73 & \cellcolor[HTML]{cdcbcb}99.62 & \textbf{55.51} & 70.67               & 71.89     \\
\multicolumn{1}{l|}{Xception}  & \multicolumn{1}{c|}{$\checkmark$}   & \multicolumn{1}{c|}{}     \hspace{0.2pt} & \textbf{89.94} & \textbf{51.91} & \cellcolor[HTML]{cdcbcb}99.36 & 47.80 & \textbf{71.73}          & \textbf{72.15}     \\
\multicolumn{1}{l|}{EfficientNet}  & \multicolumn{1}{c|}{$\times$}   & \multicolumn{1}{c|}{}   \hspace{0.2pt} & 86.43 & 54.51 & \cellcolor[HTML]{cdcbcb}99.69 & \textbf{53.61} & 64.76                              & 71.80     \\
\multicolumn{1}{l|}{EfficientNet}  & \multicolumn{1}{c|}{$\checkmark$}   & \multicolumn{1}{c|}{} \hspace{0.2pt} & \textbf{88.33} & \textbf{56.93} & \cellcolor[HTML]{cdcbcb}99.51 & 50.43 & \textbf{70.59}          & \textbf{73.16}     \\ 
\multicolumn{1}{l|}{VGG}  & \multicolumn{1}{c|}{$\times$}   & \multicolumn{1}{c|}{}  \hspace{0.2pt} & \textbf{80.33}          & 32.77          & \cellcolor[HTML]{cdcbcb}99.40 & 54.77   & 54.20                    & 64.29     \\
\multicolumn{1}{l|}{VGG}  & \multicolumn{1}{c|}{$\checkmark$}   & \multicolumn{1}{c|}{} \hspace{0.2pt}    & 76.19          & \textbf{55.32} & \cellcolor[HTML]{cdcbcb}99.10 & \textbf{55.51}   & \textbf{55.57}     & \textbf{68.34}     \\
\multicolumn{1}{l|}{F$^3$-Net}  & \multicolumn{1}{c|}{$\times$}   & \multicolumn{1}{c|}{} \hspace{0.2pt}   & 78.27          & 66.09          & \cellcolor[HTML]{cdcbcb}99.21 & \textbf{49.96}          & 60.81      & 70.86     \\   
\multicolumn{1}{l|}{F$^3$-Net}  & \multicolumn{1}{c|}{$\checkmark$}   & \multicolumn{1}{c|}{} \hspace{0.2pt}   & \textbf{79.58} & \textbf{66.25} & \cellcolor[HTML]{cdcbcb}98.89 & 44.02          & \textbf{65.76}  & \textbf{70.90}    \\
\multicolumn{1}{l|}{ResNet-50}  & \multicolumn{1}{c|}{$\times$}   & \multicolumn{1}{c|}{} \hspace{0.2pt}    & \textbf{85.14}   & 41.36          & \cellcolor[HTML]{cdcbcb}99.05 & 53.68 & \textbf{72.66}            & 70.37   \\   
\multicolumn{1}{l|}{ResNet-50}  & \multicolumn{1}{c|}{$\checkmark$}   & \multicolumn{1}{c|}{} \hspace{0.2pt}  & 84.80     & \textbf{49.86} & \cellcolor[HTML]{cdcbcb}99.30 & \textbf{62.28}          & 71.25        & \textbf{73.49}    \\ \midrule

\multicolumn{1}{l|}{Xception}  & \multicolumn{1}{c|}{$\times$}  & \multicolumn{1}{c|}{\multirow{10}{*}{Fsh}} \hspace{0.2pt}  & 71.69 & 37.63 & \textbf{48.38} & \cellcolor[HTML]{cdcbcb}99.75 & 59.94                         & 63.47   \\
\multicolumn{1}{l|}{Xception}  & \multicolumn{1}{c|}{$\checkmark$}   & \multicolumn{1}{c|}{} \hspace{0.2pt} & \textbf{72.89} & \textbf{39.60} & 47.39 & \cellcolor[HTML]{cdcbcb}99.61 & \textbf{60.07}                        & \textbf{63.91}  \\
\multicolumn{1}{l|}{EfficientNet}  & \multicolumn{1}{c|}{$\times$}   & \multicolumn{1}{c|}{} \hspace{0.2pt}    & 72.37 & 44.45 & \textbf{53.55} & \cellcolor[HTML]{cdcbcb}99.38 & 62.70                                       & 66.49   \\
\multicolumn{1}{l|}{EfficientNet}  & \multicolumn{1}{c|}{$\checkmark$}   & \multicolumn{1}{c|}{} \hspace{0.2pt}  & \textbf{72.99} & \textbf{48.22} & 50.89 & \cellcolor[HTML]{cdcbcb}99.30 & \textbf{63.19}                   & \textbf{66.91}   \\
\multicolumn{1}{l|}{VGG}  & \multicolumn{1}{c|}{$\times$}   & \multicolumn{1}{c|}{} \hspace{0.2pt}      & 67.57          & 33.94          & \textbf{50.32} & \cellcolor[HTML]{cdcbcb}98.93 & 62.35                            & 62.62   \\
\multicolumn{1}{l|}{VGG}  & \multicolumn{1}{c|}{$\checkmark$}   & \multicolumn{1}{c|}{} \hspace{0.2pt}                 & \textbf{67.76} & \textbf{44.05} & 46.51          & \cellcolor[HTML]{cdcbcb}99.22 & \textbf{63.71}    & \textbf{64.25}    \\
\multicolumn{1}{l|}{F$^3$-Net}  & \multicolumn{1}{c|}{$\times$}   & \multicolumn{1}{c|}{} \hspace{0.2pt}                & 69.04          & 36.89          & 42.30                   & \cellcolor[HTML]{cdcbcb}99.42 & 60.12   & 61.55   \\
\multicolumn{1}{l|}{F$^3$-Net}  & \multicolumn{1}{c|}{$\checkmark$}   & \multicolumn{1}{c|}{} \hspace{0.2pt}            & \textbf{71.17} & \textbf{40.48} & \textbf{45.41}    & \cellcolor[HTML]{cdcbcb}99.43 & \textbf{61.17}& \textbf{63.53}   \\
\multicolumn{1}{l|}{ResNet-50}  & \multicolumn{1}{c|}{$\times$}   & \multicolumn{1}{c|}{} \hspace{0.2pt}                & \textbf{68.48}          & 42.20          & 52.71  & \cellcolor[HTML]{cdcbcb}99.45 & \textbf{67.69}  & 66.10    \\
\multicolumn{1}{l|}{ResNet-50}  & \multicolumn{1}{c|}{$\checkmark$}   & \multicolumn{1}{c|}{} \hspace{0.2pt}            & 66.31          & \textbf{49.54} & \textbf{53.02}          & \cellcolor[HTML]{cdcbcb}99.04 & 65.87   & \textbf{66.75}    \\\midrule
                                
\multicolumn{1}{l|}{Xception}  & \multicolumn{1}{c|}{$\times$}  & \multicolumn{1}{c|}{\multirow{10}{*}{NT}} \hspace{0.2pt}    & 87.69 & 39.95 & 70.02 & 67.25 & \cellcolor[HTML]{cdcbcb}98.30                                                 &72.64  \\
\multicolumn{1}{l|}{Xception}  & \multicolumn{1}{c|}{$\checkmark$}   & \multicolumn{1}{c|}{} \hspace{0.2pt} & \textbf{89.27} & \textbf{42.25} & \textbf{71.44} & \textbf{70.79} & \cellcolor[HTML]{cdcbcb}97.24                               &\textbf{74.19}  \\
\multicolumn{1}{l|}{EfficientNet}  & \multicolumn{1}{c|}{$\times$}   & \multicolumn{1}{c|}{} \hspace{0.2pt}   & 88.24 & 40.45 & 70.99 & 72.99 & \cellcolor[HTML]{cdcbcb}98.10                                                                 &74.15  \\
\multicolumn{1}{l|}{EfficientNet}  & \multicolumn{1}{c|}{$\checkmark$}   & \multicolumn{1}{c|}{} \hspace{0.2pt}  & \textbf{89.01} & \textbf{41.04} & \textbf{75.35} & \textbf{73.81} & \cellcolor[HTML]{cdcbcb}97.92                          &\textbf{75.42}  \\ 
\multicolumn{1}{l|}{VGG}  & \multicolumn{1}{c|}{$\times$}      & \multicolumn{1}{c|}{}  \hspace{0.2pt} & 83.84          & \textbf{40.74}          & 65.86          & 67.66          & \cellcolor[HTML]{cdcbcb}96.91                           &71.00  \\
\multicolumn{1}{l|}{VGG}  & \multicolumn{1}{c|}{$\checkmark$}   & \multicolumn{1}{c|}{} \hspace{0.2pt}                 & \textbf{88.00} & 38.78                   \textbf{}& \textbf{65.94} & \textbf{68.89} & \cellcolor[HTML]{cdcbcb}94.52  &\textbf{71.22}  \\
\multicolumn{1}{l|}{F$^3$-Net}  & \multicolumn{1}{c|}{$\times$}   & \multicolumn{1}{c|}{} \hspace{0.2pt}                & 84.45          & 39.68          & 70.29          & 67.42          & \cellcolor[HTML]{cdcbcb}96.97                   &71.76  \\
\multicolumn{1}{l|}{F$^3$-Net}  & \multicolumn{1}{c|}{$\checkmark$}   & \multicolumn{1}{c|}{} \hspace{0.2pt}            & \textbf{85.86} & \textbf{44.42} & \textbf{70.85} & \textbf{71.07} & \cellcolor[HTML]{cdcbcb}96.59                   &\textbf{73.75} \\ 
\multicolumn{1}{l|}{ResNet-50}  & \multicolumn{1}{c|}{$\times$}   & \multicolumn{1}{c|}{} \hspace{0.2pt}                & \textbf{80.81}          & 45.78          & 66.52          & \textbf{74.19}          & \cellcolor[HTML]{cdcbcb}97.48 &72.95  \\
\multicolumn{1}{l|}{ResNet-50}  & \multicolumn{1}{c|}{$\checkmark$}   & \multicolumn{1}{c|}{} \hspace{0.2pt}            & 79.52                   & \textbf{47.75} & \textbf{67.13} & 74.01 & \cellcolor[HTML]{cdcbcb}{97.33}                 &\textbf{73.14}  \\ 
\midrule \bottomrule
\end{tabular}
}
\label{Tab:subset}
\end{table}

\noindent\textbf{Comparison with SOTAs.} 
We further compared the generalization ability of our method with several SOTA deepfake detectors. Specifically, we categorized the compared methods into two groups based on whether they use additional features, e.g., facial details~\cite{On_the_Detection_2020} or modalities~\cite{F3Net}, outside the backbone.
From the overall performance in Table~\ref{SOTA}, models using augmentation usually outperform the part without augmentation. 

Nonetheless, the performance of the vanilla backbones is merely competitive or lower than the SOTA methods such as CViT and Capsule. With the help of our PRLE, these backbones can mostly surpass other baselines by a large margin. 
For example, EfficientNet attains a 9\% improvement regarding the average AUC and F$^3$-Net gains 15\% AUC in DF-1.0. 
Moreover, when compared with the baselines that also use Xception as the backbone (apple-to-apple comparison), e.g., SRM, and RFM, our method shows more performance improvements of 4\% and 3\%, respectively.

\subsubsection{Comparison on FF++ subsets} 
\begin{figure}[t]
    \centering
    \includegraphics[width=0.48\textwidth]{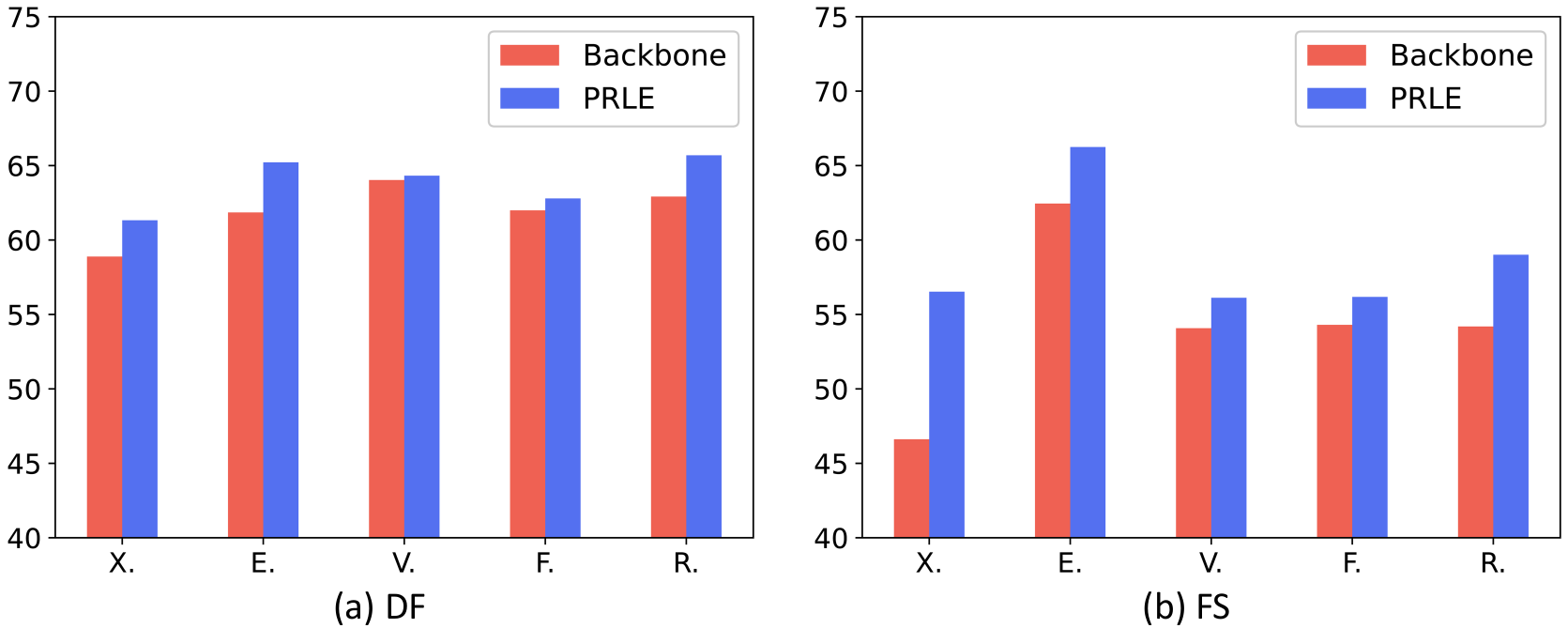}
    \caption{
    The average AUC (\%) of models trained on (a) DF and (b) FS and tested across three testing sets. $x$-axis represents five employed backbones: Xception (X.), EfficientNet (E.), VGG (V.), F$^3$-Net (F.), and ResNet-50 (R.).}
    \label{fig:AVG_subset}
    \vspace{-1em}
\end{figure}

Following previous studies~\cite{x-ray, Face_Reconstruction}, we conducted a fine-grained cross-manipulation evaluation using the FF++ dataset. Specifically, we trained the models on a single manipulation technique and tested them on the remaining techniques. The results presented in Table~\ref{Tab:subset} demonstrate that our method outperforms the five backbones on most occasions.
With the assistance of PRLE, Xception trained on DF shows 6\% and 5\% improvements in F2F and Fsh, respectively. Additionally, ResNet-50 trained on FS exhibits 12\% improvements in both DF and F2F. These results effectively validate the efficacy of our method in this cross-manipulation scenario.
We also evaluated these models on three testing datasets. 
As shown in Figure~\ref{fig:AVG_subset}, it can be observed that when trained on DF and FS, all five backbones achieve a significant improvement in average AUC scores. This implies that even under resource-constrained conditions, PRLE can effectively enhance models' generalization capability.

\subsection{Ablation Studies}
\label{abl}

\begin{table}[t]
\centering
\caption{AUC (\%) comparison of backbones with different primary region masks.}
\scalebox{0.87}{
\begin{tabular}{lccc}
\toprule \midrule
\multicolumn{1}{l|}{Method}     & DFDC & DF-1.0  & Celeb-DF \\ \midrule 
\rowcolor[HTML]{F5F4E9}\multicolumn{1}{l|}{Xception \hspace{7pt} + Primary Region}  & \textbf{69.38}           & \textbf{66.62}               & \textbf{65.94} \\       
\multicolumn{1}{l|}{\hspace{39pt} + Random}      & 66.56(\textcolor{blue}{-2.82})             & 62.38(\textcolor{blue}{-4.24})                & 61.15(\textcolor{blue}{-4.79})         \\
\multicolumn{1}{l|}{\hspace{39pt} + Landmark}    & 68.96(\textcolor{blue}{-0.42})             & 59.15(\textcolor{blue}{-7.47})                & 62.12(\textcolor{blue}{-3.82})         \\
\midrule 
\rowcolor[HTML]{F5F4E9}\multicolumn{1}{l|}{EfficientNet + Primary Region}  & \textbf{69.64}           & \textbf{74.72}               & \textbf{70.67} \\       
\multicolumn{1}{l|}{\hspace{39pt} + Random}      & 67.85(\textcolor{blue}{-1.79})             & 59.08(\textcolor{blue}{-15.44})                & 65.98(\textcolor{blue}{-4.69})         \\
\multicolumn{1}{l|}{\hspace{39pt} + Landmark}    & 68.22(\textcolor{blue}{-1.42})             & 67.97(\textcolor{blue}{-6.75})                 & 66.76(\textcolor{blue}{-3.91})         \\
\midrule 
\rowcolor[HTML]{F5F4E9}\multicolumn{1}{l|}{VGG \hspace{19pt} + Primary Region}  & \textbf{62.54}       & \textbf{68.73}               & \textbf{68.77} \\       
\multicolumn{1}{l|}{\hspace{39pt} + Random}      &59.36(\textcolor{blue}{-3.18})             & 65.97(\textcolor{blue}{-2.76})                &64.39(\textcolor{blue}{-4.38})         \\
\multicolumn{1}{l|}{\hspace{39pt} + Landmark}    &61.40(\textcolor{blue}{-1.14})             & 66.95(\textcolor{blue}{-1.78})                 &65.41(\textcolor{blue}{-3.36})         \\
\midrule 
\rowcolor[HTML]{F5F4E9}\multicolumn{1}{l|}{F$^3$-Net \hspace{13.5pt} + Primary Region}  & \textbf{69.47}           & \textbf{74.30}               & \textbf{68.17} \\       
\multicolumn{1}{l|}{\hspace{39pt} + Random}      & 64.58(\textcolor{blue}{-4.89})             & 71.29(\textcolor{blue}{-3.01})                & 64.22(\textcolor{blue}{-3.95})         \\
\multicolumn{1}{l|}{\hspace{39pt} + Landmark}    & 64.46(\textcolor{blue}{-5.01})             & 71.30(\textcolor{blue}{-3.00})                 & 64.54(\textcolor{blue}{-3.63})         \\
\midrule 
\rowcolor[HTML]{F5F4E9}\multicolumn{1}{l|}{ResNet-50 \hspace{2.2pt} + Primary Region}  & \textbf{66.02}             & \textbf{70.52}              & \textbf{61.26}           \\ 
\multicolumn{1}{l|}{\hspace{39pt} + Random}          & 65.08(\textcolor{blue}{-0.94})             & 63.37(\textcolor{blue}{-7.15})                & 59.27(\textcolor{blue}{-1.99})         \\
\multicolumn{1}{l|}{\hspace{39pt} + Landmark}        & 62.72(\textcolor{blue}{-3.30})             & 63.71(\textcolor{blue}{-7.81})                & 59.95(\textcolor{blue}{-1.31})                     \\ \midrule \bottomrule
\end{tabular}}
\label{Tab:extend_mask}
\end{table}

\begin{table}[t]
\centering
\caption{AUC (\%) comparison of backbones when applying different fusion strategies.}
\scalebox{0.87}{
\begin{tabular}{lccc}
\toprule \midrule
\multicolumn{1}{l|}{Method}     & DFDC & DF-1.0  & Celeb-DF \\ \midrule 
\rowcolor[HTML]{F5F4E9}\multicolumn{1}{l|}{Xception \hspace{7pt} + Neighboring }       & \textbf{69.38}             & \textbf{66.62}              & \textbf{65.94}           \\ 
\multicolumn{1}{l|}{\hspace{39pt} + Average (low $\tau_1$)}    & 67.07(\textcolor{blue}{-2.31})    & 66.28(\textcolor{blue}{-0.34})       & 64.46(\textcolor{blue}{-1.48})   \\
\multicolumn{1}{l|}{\hspace{39pt} + Average (high $\tau_1$)}   & 64.85(\textcolor{blue}{-4.53})    & 65.11(\textcolor{blue}{-1.51})        & 63.57(\textcolor{blue}{-2.37})    \\    
\midrule 
\rowcolor[HTML]{F5F4E9}\multicolumn{1}{l|}{EfficientNet + Neighboring}    & \textbf{69.64}           & \textbf{74.72}               & \textbf{70.67} \\      
\multicolumn{1}{l|}{\hspace{39pt} + Average (low $\tau_1$)}    & 68.90(\textcolor{blue}{-0.74})        & 72.27(\textcolor{blue}{-2.45})      & 69.34(\textcolor{blue}{-1.33})  \\
\multicolumn{1}{l|}{\hspace{39pt} + Average (high $\tau_1$)}   & 67.31(\textcolor{blue}{-2.63})        & 69.94(\textcolor{blue}{-4.78})      & 69.18(\textcolor{blue}{-1.49})  \\ 
\midrule 
\rowcolor[HTML]{F5F4E9}\multicolumn{1}{l|}{VGG \hspace{19pt} + Neighboring }       & \textbf{62.54}           & \textbf{68.73}               & \textbf{68.77} \\ 
\multicolumn{1}{l|}{\hspace{39pt} + Average (low $\tau_1$)}    & 60.82(\textcolor{blue}{-1.72})    & 65.56(\textcolor{blue}{-3.17})        & 64.47(\textcolor{blue}{-4.30})   \\
\multicolumn{1}{l|}{\hspace{39pt} + Average (high $\tau_1$)}   & 60.72(\textcolor{blue}{-1.82})    & 65.51(\textcolor{blue}{-3.22})        & 62.18(\textcolor{blue}{-6.59})    \\    
\midrule 
\rowcolor[HTML]{F5F4E9}\multicolumn{1}{l|}{F$^3$-Net \hspace{13pt} + Neighboring }       & \textbf{69.47}           & \textbf{74.30}               & \textbf{68.17} \\    
\multicolumn{1}{l|}{\hspace{39pt} + Average (low $\tau_1$)}    & 67.82(\textcolor{blue}{-1.65})    & 70.40(\textcolor{blue}{-3.90})               & 64.67(\textcolor{blue}{-3.50})   \\
\multicolumn{1}{l|}{\hspace{39pt} + Average (high $\tau_1$)}   & 66.69(\textcolor{blue}{-2.78})    & 70.21(\textcolor{blue}{-4.09})        & 64.37(\textcolor{blue}{-3.80})    \\    
\midrule 
\rowcolor[HTML]{F5F4E9}\multicolumn{1}{l|}{ResNet-50 \hspace{2.2pt} + Neighboring }       & \textbf{66.02}             & \textbf{70.52}              & \textbf{61.26}           \\ 
\multicolumn{1}{l|}{\hspace{39pt} + Average (low $\tau_1$)}    & 65.45(\textcolor{blue}{-0.57})    & 66.23(\textcolor{blue}{-4.29})       & 61.04(\textcolor{blue}{-0.22})   \\
\multicolumn{1}{l|}{\hspace{39pt} + Average (high $\tau_1$)}   & 64.93(\textcolor{blue}{-1.09})    & 62.99(\textcolor{blue}{-7.53})        & 60.82(\textcolor{blue}{-0.44})    \\    
\midrule 
\bottomrule
\end{tabular}}
\label{Tab:avg}
\end{table}

\begin{table}[t]
\centering
\caption{AUC (\%) comparison of EfficientNet when adopting masks from a single backbone, namely, Xception (X.), EfficientNet (E.), and VGG (V.). Training with the individual masks will result in a performance decrease.}
\scalebox{0.87}{
\begin{tabular}{ccc|ccc}
\toprule \midrule
\multicolumn{3}{c|}{Mask} & \multicolumn{3}{c}{Dataset} \\ \midrule
X.      & E.     & V.     & DFDC   & DF-1.0  & Celeb-DF  \\ \midrule
\rowcolor[HTML]{F5F4E9}$\checkmark$       & $\checkmark$      & $\checkmark$      & \textbf{69.64}       & \textbf{74.72}     & \textbf{70.67}           \\ \midrule  
$\checkmark$   & $\times$   & $\times$    & 68.22(\textcolor{blue}{-1.42})             & 70.83(\textcolor{blue}{-3.89})                & 67.07(\textcolor{blue}{-3.60})           \\
$\times$     &  $\checkmark$   & $\times$  & 63.69(\textcolor{blue}{-5.95})             & 72.28(\textcolor{blue}{-2.44})                 & 68.91(\textcolor{blue}{-1.76})           \\ 
$\times$    &   $\times$  & $\checkmark$   & 68.31(\textcolor{blue}{-1.33})             & 67.57(\textcolor{blue}{-7.15})                 & 64.22(\textcolor{blue}{-6.45})                     \\ \midrule \bottomrule
\end{tabular}}
\label{Tab:masks}
\end{table}

\begin{figure}[t]
    \centering
    \includegraphics[width=0.48\textwidth]{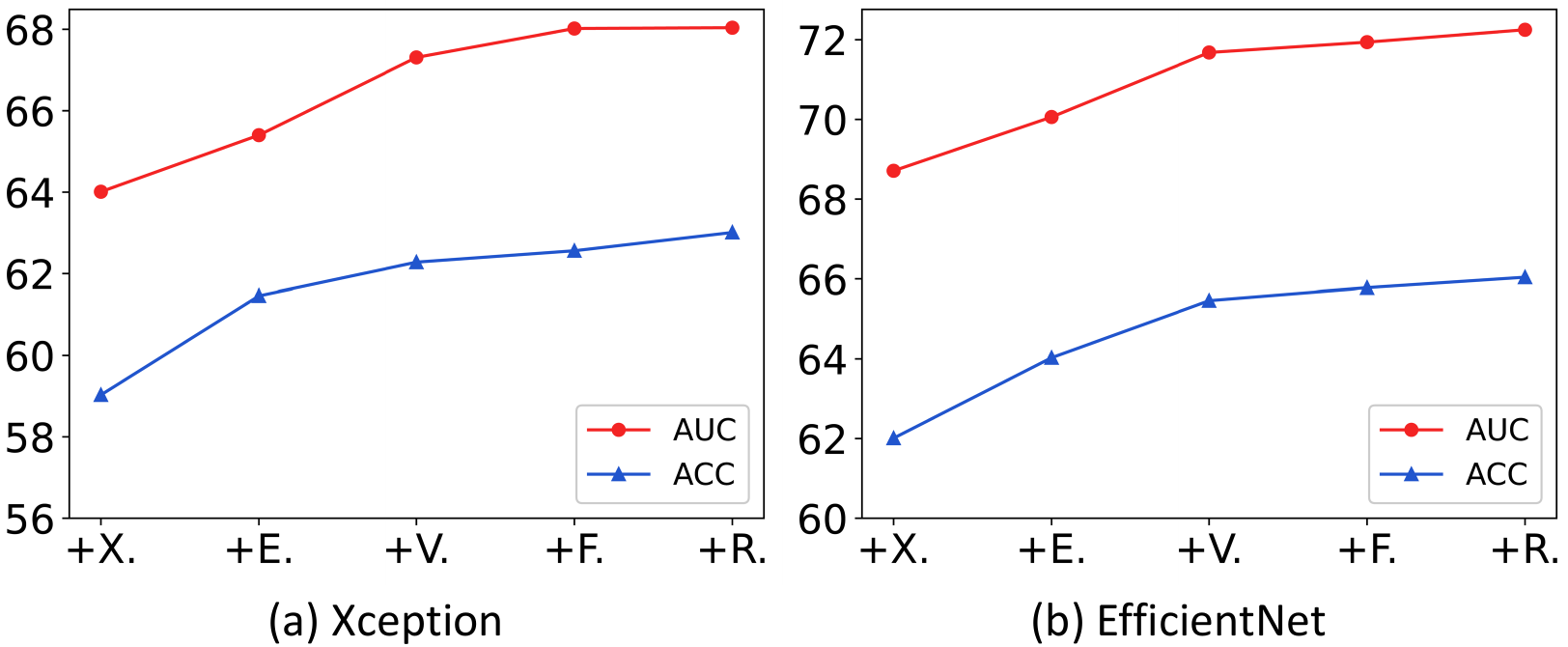}
    \caption{Performance (\%) of (a) Xception and (b) EfficientNet when utilizing attention maps obtained from different detectors. $x$-axis - the sequential addition of attention maps from detectors Xception (+X.), EfficientNet (+E.), VGG (+V.), F$^3$-Net (+F.), and ResNet-50 (+R.) into the fusion strategy.}
    \label{fig:different_detectors}
\end{figure}

\begin{table}[t]
\centering
\caption{AUC (\%) comparison of EfficientNet and ResNet-50 backbones with the removal of the dynamic exploitation.}
\scalebox{0.87}{
\begin{tabular}{lccc}
\toprule \midrule
\multicolumn{1}{l|}{Method}     & DFDC & DF-1.0  & Celeb-DF \\ \midrule 
\rowcolor[HTML]{F5F4E9}\multicolumn{1}{l|}{EfficientNet}    & \textbf{69.64}           & \textbf{74.72}               & \textbf{70.67} \\      
\multicolumn{1}{l|}{~~\textit{w/o} exploitation}      & 68.86(\textcolor{blue}{-0.78})             & 70.42(\textcolor{blue}{-4.30})                & 65.45(\textcolor{blue}{-5.22})         \\  \midrule 
\rowcolor[HTML]{F5F4E9}\multicolumn{1}{l|}{ResNet-50}         & \textbf{66.02}             & \textbf{70.52}              & \textbf{61.26}           \\ 
\multicolumn{1}{l|}{~~\textit{w/o} exploitation}       & 65.86(\textcolor{blue}{-0.16})             & 63.32(\textcolor{blue}{-7.20})                 & 60.83(\textcolor{blue}{-0.43})                     \\ \midrule \bottomrule
\end{tabular}}
\label{Tab:refine}
\end{table}

\subsubsection{Comparison on alternative region masks}
In contrast to the proposed PRLE method, which utilizes attention maps to locate primary regions, we employed two alternative approaches to generate masks: i) \textbf{Random Selection Approach}, where masks are derived from randomly selected rectangular regions within the image, and ii) \textbf{Landmark-based Approach}, where facial regions indicated by landmarks, such as the mouth or nose, are randomly selected and masked.

The model performance with these alternative masks is summarized in Table~\ref{Tab:extend_mask}. From the table, it can be observed that our proposed primary region masks outperform both randomly generated and landmark-based masks by a significant margin. For instance, on DF-1.0, the utilization of alternative masks resulted in a maximum performance decrease of 15\%. Besides, on Celeb-DF, the average decline in performance is approximately 2.5\%. This discrepancy can be attributed to the fact that neither of the alternative masks explicitly captures the primary evidence used by models to distinguish between real and fake images, potentially rendering them ineffective in mitigating overfitting. Among these two masks, the landmark-based masks perform better since randomly generated masks may fail to cover any facial regions, thus not contributing to the detection of real and fake faces.

\subsubsection{Comparison on strategies for primary region localization} 
As described in Section~\ref{sec:method}, the primary regions maps are fused with the neighboring fusion strategy from the attention maps of three deepfake detectors, i.e., Xception, EfficientNet, and VGG. We therefore conducted experiments to verify the effectiveness of this fusion approach.

\noindent\textbf{Masks from different fusion strategies.}
We applied different fusion strategies to evaluate our PRLE. Specifically, we employed two average fusion methods with different $\tau_1$, as depicted in Equation~\ref{eqn:tau1_mask}, as alternatives to the proposed neighboring fusion approach. The summarized results are reported in Table~\ref{Tab:avg}.
Notably, the models using average strategies perform worse compared to our PRLE. For instance, Xception experiences a decrease of 4.5\% on DFDC, while EfficientNet decreases by 4.7\% on DF-1.0. An intuitive explanation can be derived from Figure~\ref{fig:compare_strategy}, where the attention maps generated by average strategies may contain excessive noise or incomplete regions. These factors can hinder the effective enhancement of model generalization.
It is also observed that a lower $\tau_1$ yields better performance compared to a higher value. This can be attributed to the dynamic exploitation stage, where we refine the mask and reduce some noise from the initial stage.

\noindent\textbf{Masks from different detectors.}
We further explored the impact of utilized detectors. 
Firstly, we employed EfficientNet as the backbone and trained it with masks from every single detector used in the proposed PRLE.
In Table~\ref{Tab:masks}, using individual masks jeopardizes the model performance. For instance, the performance decreases by around 7\% when trained with the masks from VGG and tested on DF-1.0. 
We attributed this result to the static localization stage integrating information from different backbones, enriching the information of the mask and alleviating the bias on a single detector.

Furthermore, we investigated the influence of the number of detectors on model generalization. 
Specifically, we progressively fuse attention maps from different detectors to obtain masks.
As shown in Figure~\ref{fig:different_detectors}, we reported the average AUC and ACC of Xception and EfficientNet with respect to different detectors. Generally, both models exhibit improvement as the number of detectors increases. For instance, the average AUCs of EfficientNet are 68.71\%, 70.06\%, 71.68\%, 71.94\%, and 72.25\% for 1 to 5 detectors, respectively. However, this growth comes at the cost of increased computation, indicating the presence of a trade-off.
Besides, this growth starts to decelerate when the number of detectors reaches three. Based on these findings, we have empirically employed three detectors, i.e., Xception, EfficientNet, and VGG, to strike a reasonable balance between performance and computational costs.

\subsubsection{Comparison on mask exploitation} 
We validated the importance of our dynamic exploitation module, which augments images with variable $\alpha$.
Specifically, we discarded the dynamic exploitation and converted the fused attention maps to binary masks directly. The results are reported in Table~\ref{Tab:refine}.
One can observe that dynamic exploitation plays a key role in improving generalizability, as it contributes 7\% improvement to ResNet-50 on DF-1.0. Besides, EfficientNet degrades around 3\% across testing sets without exploitation.

\begin{figure*}[t]
    \centering
        \includegraphics[width=0.95\textwidth]{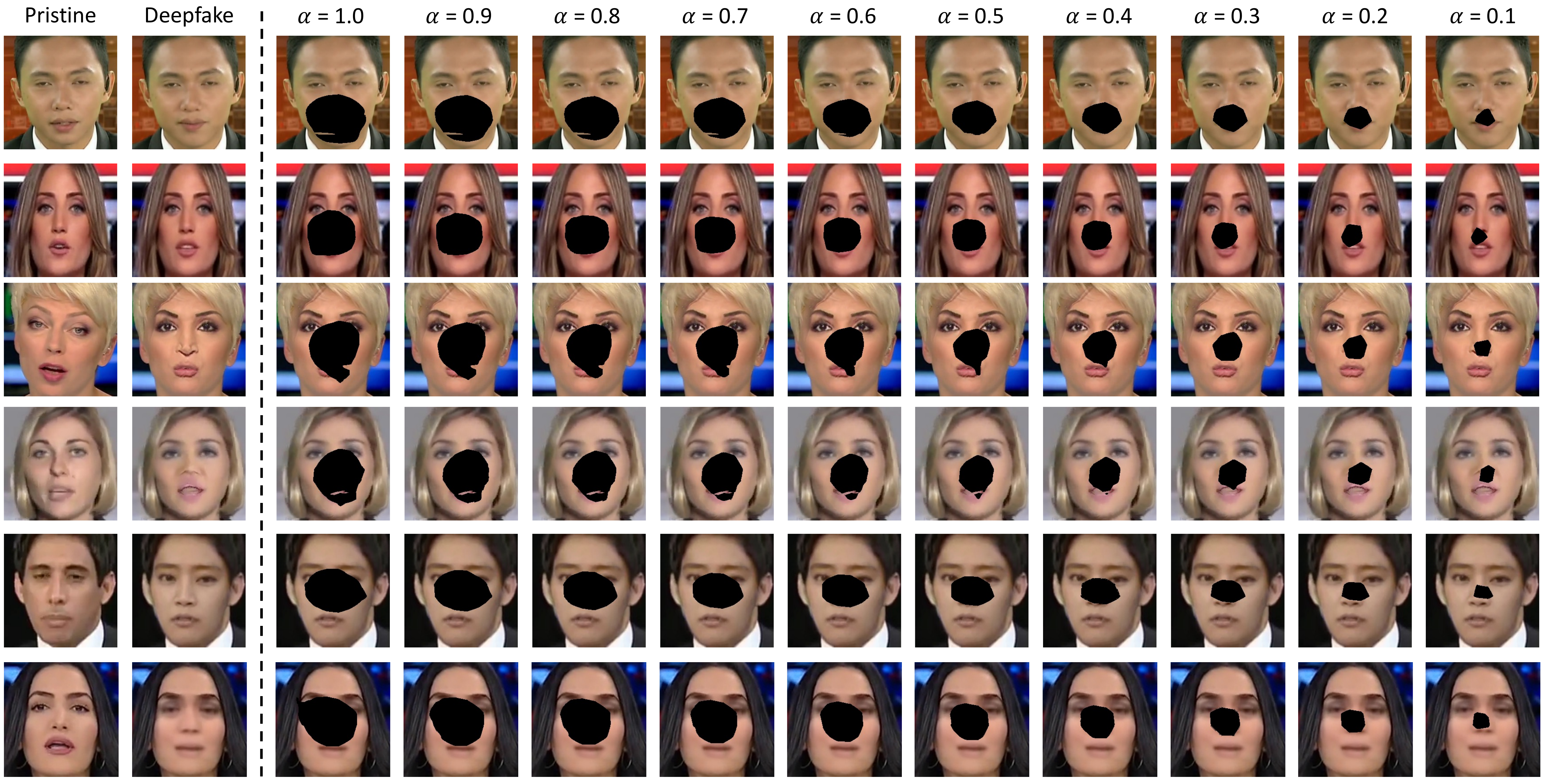}
    \caption{Visualization of masks with $\alpha$ ranging from 1.0 to 0.1.}
    \label{fig:mask_alpha} 
\end{figure*}

\begin{figure*}[t]
    \centering
        \includegraphics[width=0.95\textwidth]{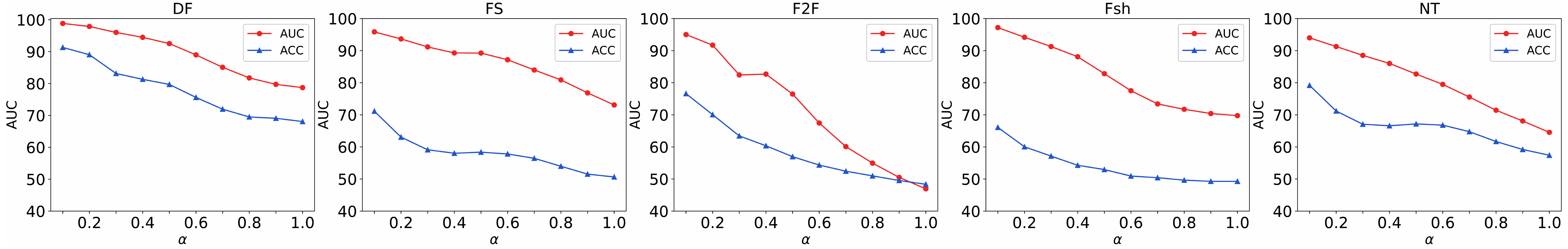}
    \caption{Performance (\%) of Xception when tested on the masked images from different FF++ subsets.}
    \label{fig:Mask_res_alpha} 
    \vspace{-0.5em}
\end{figure*}

\subsection{Qualititive Studies}
\label{qualititive}
\subsubsection{Mask ratio $\alpha$} 
We studied the impact of different $\alpha$ values and visualized the corresponding masks in Figure~\ref{fig:mask_alpha}. The visualization demonstrates that as $\alpha$ increases, the masked regions expand from a central point toward the peripheral areas.
For instance, cases 1 and 2 primarily focus on the \textbf{lips} region, while cases 3 and 4 highlight the \textbf{nose}. Cases 5 and 6 concentrate on the \textbf{cheek} and \textbf{eyes}. This variation in masked regions is attributed to the fact that different forgery methods tend to target specific regions. For example, cases 1 and 2 are from the NT subset of FF++, whose main modification involves mouth movements, resulting in detectors placing significant attention on the lips.

We further investigated the performance change of Xception using different $\alpha$ in Figure~\ref{fig:Mask_res_alpha}. Specifically, the model is trained only on raw images from a specific subset and tested on the corresponding masked images. We can see that the model performance drops with the increase of $\alpha$ and then converges at around 50-70\%. 
This suggests that the masking strategy gradually removes the key clues that the model depends on to determine authenticity.
It is important to clarify that the primary regions describe the areas that detectors utilize to determine authenticity. However, these regions may not encompass the complete regions of all manipulated artifacts. 

\begin{figure}[t]
\centering
\includegraphics[width=0.48\textwidth]{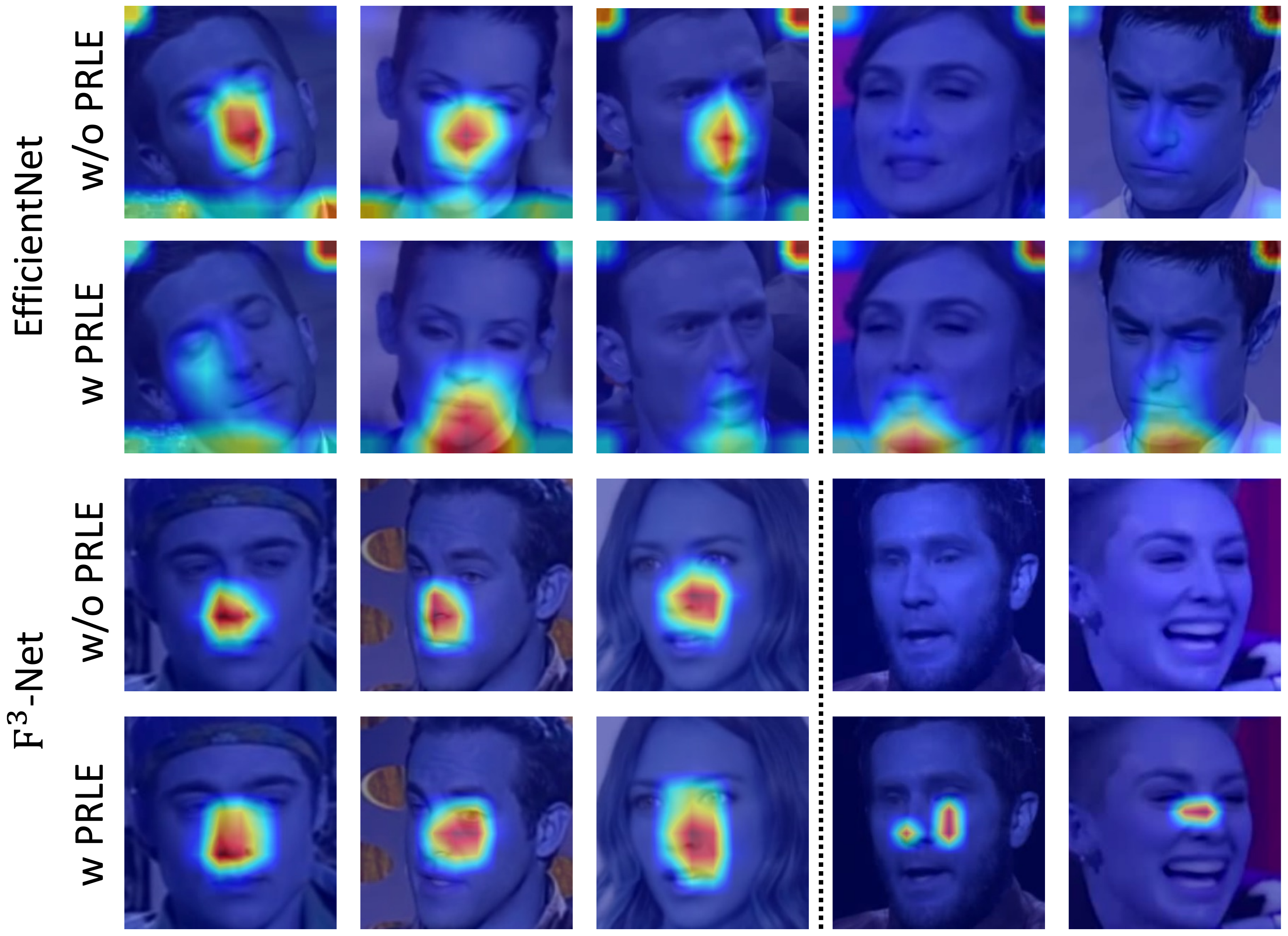}
\caption{Non-cherry-picked heat maps on Celeb-DF. The first three columns show the images that both the vanilla backbones and PRLE can detect accurately, while the images in the last two columns are those that the backbones fail to distinguish.}
\label{fig:heatmap_compare}
\end{figure}

\subsubsection{Heat maps}
To qualitatively evaluate our method, we employed two backbones, i.e., EfficientNet and F$^3$-Net, to demonstrate several heat maps under the cross-dataset setting in Figure~\ref{fig:heatmap_compare}.
We can observe that for both vanilla backbones, the attention regions are concentrated on the nose and upper lip, indicating that these two are overfitted on the training set FF++. Compared with the backbones, PRLE focuses on broader face regions, such as eyes and chins, showing that more cues beyond the primary regions are utilized for detection. 
Moreover, the last two columns illustrate the forged images that the backbones cannot distinguish from the real ones. The attention maps show that the backbones fail to leverage accurate clues, and thus lead to wrong predictions. In contrast, PRLE predicts correctly by exploring more regions.

\begin{figure}[t]
\centering
\includegraphics[width=0.48\textwidth]{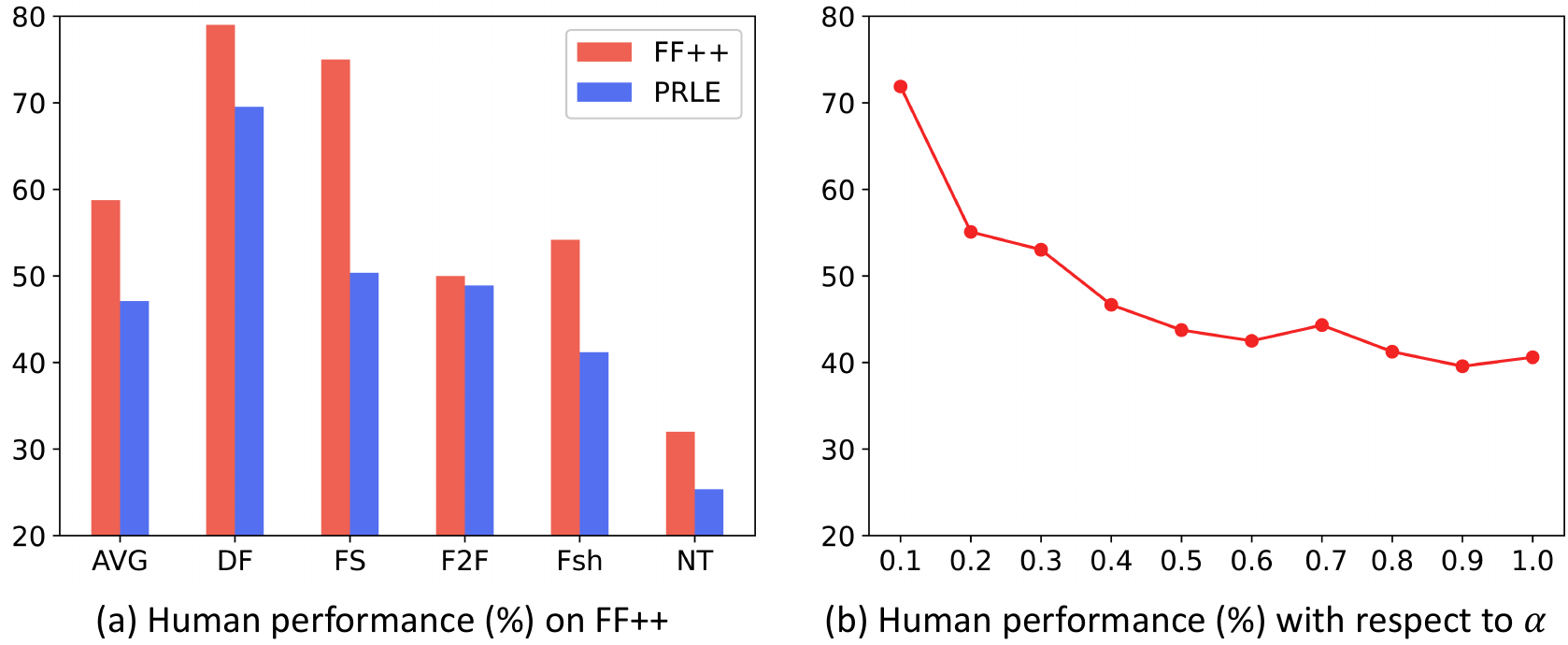}
\caption{User study results. (a) Deepfake detection results of the original FF++ dataset and the FF++ dataset with the primary region augmentation for different sub-datasets. (b) Results with respect to different $\alpha$.}
\label{fig:human_study}
\end{figure}

\subsection{User Study}
\label{Sec: User Study}
We performed a detailed user study which includes 86 participants in total. Subjects are asked to classify the deepfake images with primary region masks, and we ensured that each identity appeared only once. Our evaluation requires each participant to take several minutes to look at 50 images, resulting in 4,300 human decisions.

The results are presented in Figure~\ref{fig:human_study}{a}. Compared to the user evaluation on FF++ without primary region masks, 
subjects' performance decreased by 10\%, indicating that the primary regions influence the real/fake judgment of images. In addition, we conducted user experiments with a fixed $\alpha$. Figure~\ref{fig:human_study}{b} illustrates that as the masked area increases, the accuracy continues to drop. 
Combining the results from these human decisions, we conclude that the primary region masks play an important role in affecting the detection judgment.

\begin{table}
\centering
\caption{Model efficiency comparison. All the models are trained and tested on two RTX 3090 GPUs with a batch size of 64.}
\scalebox{0.87}{
\begin{tabular}{lcccc}
\toprule
\midrule
\multicolumn{1}{l|}{\multirow{2}{*}{Model}} & \multirow{2}{*}{GFLOPs} & \multirow{2}{*}{Param.} & \multicolumn{2}{c}{Throughput} \\ \cmidrule{4-5} 
\multicolumn{1}{l|}{}                       &                        &                         & Train.     & Infer.    \\ \midrule
\multicolumn{1}{l|}{Xception}             & 540.15  & 20.81M      & 269.72    &  902.98  \\
\multicolumn{1}{l|}{Xception + PRLE}      & 540.15  & 20.81M      & 262.42    &  902.98  \\ \midrule
\multicolumn{1}{l|}{ResNet-50 }           & 493.38  & 25.56M      & 281.19    &  916.58  \\  
\multicolumn{1}{l|}{ResNet-50 + PRLE }    & 493.38  & 25.56M      & 270.30    &  916.58  \\ \midrule
\bottomrule
\end{tabular}
}
\label{Tab: efficiency}
\end{table}

\subsection{Efficiency Analysis}
\label{Sec:effi}
Table~\ref{Tab: efficiency} illustrates the efficiency impact of our PRLE method. We compared the computational complexity (Giga floating point operations, GFLOPs), model parameters, and video throughput (videos/s) during training and inference of the Xception and ResNet-50 backbones.
It can be observed that PRLE introduces only negligible computational time to the original detectors. For instance, incorporating PRLE into Xception incurs less than a 3\% additional time cost. This marginal decrease in speed primarily results from the refinement of masks in the dynamic exploitation stage. Considering the generalization improvement brought about by PRLE, this slight time overhead is deemed acceptable.
Moreover, apart from a minor computational cost during data processing, PRLE does not affect the computational complexity and parameters of the backbone model, thus maintaining the same inference speed as the model without PRLE. 
Admittedly, there is a small one-time upfront cost associated with integrating attention maps into primary region masks during the static localization stage. However, this cost is relatively minor. In our experiments, this stage consumes ten images per second.

\section{Conclusion and Discussion}
Deepfake detectors tend to overfit primary regions, limiting their generalization to unseen data or algorithms. In this work, we address this challenge from a novel regularization view and propose an effective data augmentation method.
Our static localization and then dynamic exploitation of primary regions strategy enables models to explore more cues for authenticity detection. Due to its plug-and-play virtue, our method is compatible with most existing deepfake detectors. Unlike conventional data augmentation approaches, our method does not enlarge the dataset size, avoiding burdening the training efficiency. 
We apply this method to several strong backbones and observe significant performance improvement in terms of generalization.
In addition to its effectiveness in generalizable deepfake detection, our data augmentation approach can also help other promising future research, such as deepfake localization and segmentation.

\ifCLASSOPTIONcaptionsoff
  \newpage
\fi

\normalem
\bibliographystyle{IEEEtran}
\bibliography{ieeebib}

\end{document}